\documentclass[10pt,twocolumn,letterpaper]{article}

\usepackage{iccv}
\usepackage{times}
\usepackage{epsfig}
\usepackage{graphicx}
\usepackage{amsmath}
\usepackage{amssymb}
\usepackage{comment}

\DeclareMathOperator*{\argmin}{arg\,min}
\usepackage{color, }
\usepackage{multirow}
\usepackage{subcaption}
\usepackage{siunitx}
\usepackage{xcolor}
\usepackage{color, colortbl}
\definecolor{rowhighlight}{gray}{0.9}
\usepackage{cuted}
\usepackage{changepage}
\usepackage{booktabs}
\usepackage{pifont}

\newcommand{\norm}[1]{\left\lVert#1\right\rVert}
\renewcommand{\vec}[1]{\mathbf{#1}}

\definecolor{training}{RGB}{103,169,207}
\definecolor{inference}{RGB}{239,138,98}

\definecolor{grey}{RGB}{80,80,80}
\definecolor{darkgrey}{RGB}{55,55,55}
\definecolor{lightgrey}{RGB}{170,170,170}

\newcommand{\cmark}{\ding{51}}
\newcommand{\xmark}{\ding{55}}

\def\meth{SimNP}

\usepackage[breaklinks=true,bookmarks=false]{hyperref}

\iccvfinalcopy 

\def\iccvPaperID{****} 

\ificcvfinal\fi

\begin{document}

\title{SimNP: Learning Self-Similarity Priors Between Neural Points}

\author{Christopher Wewer\textsuperscript{1}\\
\and
Eddy Ilg\textsuperscript{2}\\
\and
Bernt Schiele\textsuperscript{1}\\
\and
Jan Eric Lenssen\textsuperscript{1}\\
\and
\textsuperscript{1}Max Planck Institute for Informatics, Saarland Informatics Campus, Germany\\
\textsuperscript{2}Saarland University, Saarland Informatics Campus, Germany\\
{\tt\small \{cwewer, jlenssen\}@mpi-inf.mpg.de}
}

\makeatletter
\let\@oldmaketitle\@maketitle
\renewcommand{\@maketitle}{\@oldmaketitle

\newenvironment{nonfloat}
  {%
   \par\nopagebreak\vspace{\medskipamount}%
   \noindent\begin{minipage}{\linewidth}
   \captionsetup[subfigure]{
     margin=0pt,font+=small,labelformat=parens,labelsep=space,
     skip=-6pt,list=false,hypcap=false
   }%
  }
  {\end{minipage}}

\begin{center}
    \begin{adjustwidth}{0pt}{0pt}
        \vspace{-20pt}
        \centering
      \begin{nonfloat}
\centering
\begin{minipage}{0.368\textwidth}
  \centering

  \includegraphics[width=\textwidth]{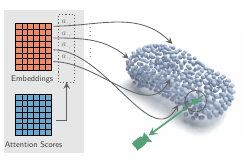}

  \textbf{a)} Neural Point Representation
\end{minipage}\hfill
\begin{minipage}{0.622\textwidth}
  \centering
  \includegraphics[width=\textwidth]{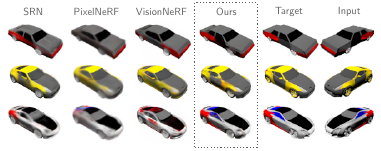}
  \textbf{b)} Single-view reconstruction comparison.
\end{minipage}
\captionof{figure}{Figure 1\textbf{a)} We present \textbf{\meth}, a renderable neural point radiance field that learns category-level self-similarities from data by connecting neural points to embeddings via optimized bipartite attention scores. \textbf{b)}  The learned self-similarities can be used to transfer details from single- or few-view observations to unobserved, similar and symmetric parts of objects.}
\label{fig:teaser}
\end{nonfloat}
    \end{adjustwidth}
\vspace{0.05cm}
\end{center}
 }
\makeatother
\maketitle
\ificcvfinal\thispagestyle{empty}\fi

\begin{abstract}
\vspace{-0.1cm}
    Existing neural field representations for 3D object reconstruction either (1) utilize object-level representations, but suffer from low-quality details due to conditioning on a global latent code,
or (2) are able to perfectly reconstruct the observations, but fail to utilize object-level prior knowledge to infer unobserved regions. %
We present \meth{}, a method to learn category-level self-similarities, which combines the advantages of both worlds by connecting neural point radiance fields with a category-level self-similarity representation. Our contribution is two-fold. (1) We design the first neural point representation on a category level by utilizing the concept of coherent point clouds. The resulting neural point radiance fields store a high level of detail for locally supported object regions. (2) We learn how information is shared between neural points in an unconstrained and unsupervised fashion, which allows to derive unobserved regions of an object during the reconstruction process from given observations. %
We show that \meth{} is able to outperform previous methods in reconstructing symmetric unseen object regions, surpassing methods that build upon category-level or pixel-aligned radiance fields, while providing semantic correspondences between instances.
\vspace{-0.5cm}
\end{abstract}

\section{Introduction}
\label{sec:introduction}
The human visual system succeeds in deriving 3D representations of objects just from incomplete 2D observations. Key to this ability is that given observations are successfully complemented by previously learned information about the 3D world. Replicating this ability has been a longstanding goal in computer vision.

Since the task of reconstructing complete objects relies on generalization from a set of known examples, deep learning is an intuitive solution. The common approach is to use large amounts of data to train a category-level model~\cite{deepsdf, srn, pixelnerf, visionnerf, pointsetgen, deformnet, dispointflow, codenerf, onet, gsm, posendf, toch} and let the reconstruction process combine observations with the prior knowledge learned from data, which we refer to as the \emph{data prior}. Notably, this introduces an inherent trade-off between contributions of the data prior and the observations. 

On one extreme of the spectrum, NeRF-like methods~\cite{nerf, nerfpp, mipnerf} do not use a data prior at all. With a high number of observations and an optimization process, they are able to nearly perfectly reconstruct novel views of scenes and objects. However, this renders them incapable of deriving unseen regions. On the other extreme of the spectrum are methods that learn the full space of radiance or signed distance functions belonging to a specific object category, such as SRN~\cite{srn} and DeepSDF~\cite{deepsdf}. While these methods succeed in learning a complete representation of objects on an abstract level, they fail to represent individual details and the reconstruction process often performs retrieval from the learned data prior~\cite{sv3dlearn}. A similar behaviour has also been observed for generative models based on GANs~\cite{eg3d} or diffusion~\cite{renderdiffusion}, which yield visually impressive results but still diverge from given observations. 

The key challenge lies in combining the strengths of the methods from both ends of the spectrum. While one can perform a highly detailed reconstruction of the visible regions, the object model should allow to reuse this information in unseen regions. To this end, it is important to know that most objects show many structured self-similarities, often arising from different types of symmetries, such as point/plane symmetries or more general variants. None of the current approaches try to explicitly learn such self-similarities to perform better inference.

This is where \meth~comes in. We propose a better data prior vs. observation trade-off by combining the best of both worlds: (1) a category-level data prior encoding self-similarities on top of a \mbox{(2) local} representation with test-time optimization. 
Instead of learning the \emph{full} space of radiance functions for a given category, we move to learning a data prior one level of abstraction higher, i.e., we learn \emph{how information can be shared} between local object regions. This enables us to learn characteristic self-similarity patterns from training data, which are used to propagate information from visible to invisible parts during inference. 

As learning a representation for category-level self-similarities implies modeling relationships between local regions of objects, a key observation in this work is that neural point representations are especially well-suited to describe such relationships. Besides their capacity to capture high-frequency patterns, the underlying sparse point cloud allows for explicit formulations of similarities. 

In summary, the contributions of our work are: 
\begin{enumerate}
\itemsep0em 
    \item We present the first generalizable neural point radiance field for representation of objects categories. 
    \item We propose a simple but effective mechanism that learns general similarities between local object regions in an unconstrained and unsupervised fashion.

    \item 
    We show that our model improves upon state of the art in reconstructing unobserved regions from a single image and by outperforming existing two-view methods by a large margin. At the same time, it is more efficient in training and rendering. 

\end{enumerate}

\section{Related Work}
\label{sec:related_work}

\paragraph{Reconstruction from Observations Only}

While 3D reconstruction has traditionally been dominated by multi-stage pipelines~\cite{schoenberger2016sfm,schoenberger2016mvs}, 
NeRF-like approaches~\cite{nerf, nerfpp, mipnerf} revolutionized novel view synthesis by using a volumetric density representation in continuous 3D space. Although such approaches can achieve very accurate reconstructions, they only work for the visible regions and require a large number of input views.

\paragraph{Reconstruction with Data Priors}

To reduce the number of input views required for the reconstruction, many ways to leverage data priors have been proposed. 
\emph{Pixel-Aligned} methods such as PixelNeRF~\cite{pixelnerf} leverage image-based rendering and use features that represent data priors obtained from CNNs~\cite{ibrnet, grf} or vision transformers~\cite{visionnerf} trained on large amounts of data. The feature for a 3D location in space is then derived from the associated local 2D image features. Methods specialized on multi-view input additionally utilize Multi-View Stereo (MVS)~\cite{mvsnerf, stereoradiance, geonerf}. FE-NVS~\cite{fe-nvs} uses an autoencoder-like architecture with a 2D and 3D U-Net.
\emph{Voxel-Based} approaches use an MLP taking in a local latent code~\cite{deepls, nsvf} to learn a data prior of geometry and appearance for a single voxel.  
\emph{Point-Based} approaches~\cite{pbgraphics, pbgraphics+, pointnerf} first obtain a point cloud from an \mbox{RGB-D} sensor or with MVS, and then initialize features for the points from CNNs similar as in image-based rendering. 
In contrast to the above, our method introduces a combined global and local representation. The local features of our approach can be optimized during test time and are propagated to unobserved regions via our learned category-specific attention.

\begin{figure*}[h]
  \centering
  \includegraphics[width=\linewidth]{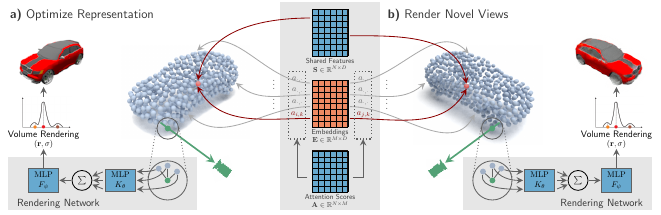}
  \caption{
    \textbf{Overview of \meth.} Our method is a category-level, coherent neural point radiance field, where points are connected to embedding vectors $\mathbf{E}$ via learnable attention scores $\mathbf{A}$. The representation can be rendered using ray marching and a neural renderer. \textbf{(a)} During training, all parameters (\textcolor{training}{$\blacksquare$}, \textcolor{inference}{$\blacksquare$}) are optimized using multi-view supervision. Networks, features~$\mathbf{S}$, and scores are shared over the category (\textcolor{training}{$\blacksquare$}), while embeddings are instance-specific (\textcolor{inference}{$\blacksquare$}). During inference, only embeddings $\mathbf{E}$ (\textcolor{inference}{$\blacksquare$}) are optimized from observations. In case of similar points $i,j$ (\eg, those shown in red), the network learned $a_{i,k} \approx a_{j,k}\,\forall\,k$ during training. Thus, supervision from one side means only one of points $i$ and $j$ needs to be visible to infer the value of embedding $k$. \textbf{(b)} Given optimized embeddings, we can render the object from novel views.}
    
  \label{fig:method}
\end{figure*}

\paragraph{Reconstruction with Object-Level Data Priors}
Early works for 3D reconstruction from single or few images leverage voxel grids~\cite{3dr2n2, predrepobj, learninglatent}. However, while being able to reconstruct complete objects, for the latter it was shown that they do not actually perform reconstruction but image classification~\cite{sv3dlearn} and are therefore on the far end of data prior in the data prior vs. observation trade-off. 

Some approaches model objects via 3D point clouds~\cite{pointsetgen, deformnet, dispointflow} but do not model surfaces and appearance. More recently, approaches that model a continuous representation of 3D space with MLPs were proposed. Scene Representation Networks~\cite{srn}~(SRN) model the scene as a continuous feature field in 3D space. Implicit representations of surfaces were introduced in DeepSDF~\cite{deepsdf} and Occupancy Networks~\cite{onet, gsm}. While these approaches can be extended to model the appearance on the surfaces~\cite{idr, dvr}, recent approaches leverage neural radiance fields~\cite{codenerf}. All of the mentioned models can be parameterized by a global latent code to model a distribution over objects as an object-level data prior. However, since prior and representation are global, the reconstructions usually lack details. In contrast, our method learns a global geometry prior with local appearance features enabling higher frequency details. We are the first to introduce a point-based neural radiance field representation on object level.

\paragraph{Modeling Self-Similarity and Symmetry}
To our knowledge, learning general self-similarities of 3D structures is a novel concept and has not been explored before. However,
many early works proposed to use or recover predefined symmetries to obtain improved reconstructions~\cite{shapefromsym, exploitsym, extractstruct, affinerecon, mirrorsym, dense3dsym, twoview, recon3Dmirror, seeingglass, autosweep}, which can be seen as a specific instance of self-similarity. More recent work performs reconstruction of a single instance by jointly optimizing the position and orientation of a symmetry plane~\cite{snes} or providing the symmetry as input~\cite{symmnerf}. Modeling plane or rotational symmetry is also common in object reconstruction in the wild~\cite{unsupsym, derendering, catmesh, unannotatedimage}. However, the constraints are only used for training and not for inference. In contrast to all of the above, SimNP learns arbitrary self-similarities from data without supervision. 

VisionNeRF~\cite{visionnerf} extends PixelNeRF~\cite{pixelnerf} with a Vision Transformer~(ViT)~\cite{vit}, which is able to globally propagate features among different rays. Therefore, one can argue that the transformer is able to learn category-level symmetries, but only implicitly, and only in 2D pixel space. 
In contrast, \meth~operates directly in 3D and learns local self-similarity relationships explicitly, which improves reconstructions of symmetric parts, as our results show.

\section{Self-Similarity Priors between Neural Points}
\label{sec:method}

In this section, we present \meth. An overview is shown in Figure~\ref{fig:method}. At the heart of our method is a neural point representation that is comprised of a point cloud with attached feature vectors $\mathcal{P}=(\mathbf{P}, \mathbf{S}, \mathbf{F})$ with point positions \mbox{$\mathbf{P}\in \mathbb{R}^{N\times 3}$} and two sets of point features \mbox{$\mathbf{S}\in \mathbb{R}^{N\times D}$} and \mbox{$\mathbf{F}\in \mathbb{R}^{N\times D}$}. The point features $\mathbf{S}$ are shared across the whole category and encode a point identity, 
while features $\mathbf{F}$ and positions $\mathbf{P}$ are individual for each instance, encoding local density and radiance. Features $\mathbf{F}$ are not explicitly stored but derived from embeddings $\mathbf{E}$ (\cf Sec.~\ref{sec:symmetry_representation}). Note that we require our point clouds to be \emph{coherent}, meaning that over different instances of one category, a single point with index $i$ describes roughly the same part of the object, \eg, the area around the right rear mirror of a car. 

\paragraph{Overview.} The neural points are connected to a large set of embeddings $\mathbf{E}$ via bipartite attention scores $\mathbf{A}$ representing our category-level self-similarity, which is further detailed in Section~\ref{sec:symmetry_representation}. The neural point cloud can be volumetrically rendered from arbitrary views by using a network to decode color and density as described in Section~\ref{sec:neural_point_representation}. We employ the autodecoder framework~\cite{deepsdf}, finding the optimal embeddings via optimization in training and inference, which is explained in Section~\ref{sec:training_and_inference}. After assuming given point clouds for the introduction of our neural point radiance field, we cover the  coherent point cloud prediction from single images independently in Section~\ref{sec:point_clouds}.

\subsection{Representing Category-Level Self-Similarity}
\label{sec:symmetry_representation}
We learn self-similarity by learning how information can be shared between coherent neural points. 
There are two characteristics why neural point clouds are a well-suited representation for learning such explicit self-similarities: (1) They disentangle \emph{where} the geometry is (represented by $\mathbf{P}$) from  \emph{how} it looks (represented by $\mathbf{F}$) and (2) they provide a discrete sampling of local parts that stays coherent across different instances due to this disentanglement. 

Since, with coherent point clouds, category-level self-similarities are invariant to variations in point location,
similarities can be formulated on top of the neural point features $\mathbf{F}$, independent of individual instances of $\mathbf{P}$. 

Formally, we store local density and radiance information in a larger number of embeddings $\mathbf{E} \in \mathbb{R}^{M \times D}$ and connect these to our neural points by
\begin{equation}
    \mathbf{F} = \textnormal{softmax}(\mathbf{A})\cdot \mathbf{E} \textnormal{,}
\end{equation}
where $\mathbf{A} \in \mathbb{R}^{N\times M}$ are learnable attention scores between $N$ points and $M$ embeddings. The matrix $\mathbf{A}$ is shared across the whole category and encodes the category-level self-similarity prior. It learns to connect two object points to the same embeddings, if they are similar and share information (see also Figure~\ref{fig:method}). Then, during reconstruction with single or few views, it is sufficient if an embedding receives gradients from only one of the connected points.

\subsection{Neural Point Rendering}
\label{sec:neural_point_representation}
Rendering a neural point cloud $\mathcal{P}=(\mathbf{P}, \mathbf{S}, \mathbf{F})$ follows a ray casting approach with a single message passing step from neural points to ray samples, similar to that of PointNeRF~\cite{pointnerf}. We cast rays through each pixel and sample points along the rays. An inherent advantage of point-based neural fields is that ray samples which are not in the vicinity of any neural points can be discarded before neural network application, which increases performance in contrast to dense approaches.

The rendering network consists of two MLPs, a kernel MLP $K_\theta$ describing local patches around each neural point, and a rendering MLP $F_\psi$ producing density and radiance from aggregated features.
Given a ray sample point $\vec{x}\in \mathbb{R}^3$, an intermediate feature vector $\vec{h}$ is computed as 
\begin{equation}
    \mathbf{h} = \frac{1}{W} \sum_{i\in \mathcal{N}(\vec{x})} w(\vec{x}, \mathbf{p}_i) \cdot K_\theta(\vec{x} - \mathbf{p}_i, \mathbf{f}_i, \mathbf{s}_i) \textnormal{,}
\end{equation}
where $\mathcal{N}(\vec{x})$ contains the indices of the $k$-nearest neural points ($k=8$ in our case) of $\vec{x}$ from $\mathbf{P}$ within a radius $r$. $w$ is a weight based on inverse point distance:
\begin{equation}
\label{eq:weights}
    w(\vec{x}, \vec{p}_i) = \frac{1}{\norm{\vec{x} - \vec{p}_i}}_2 
\end{equation}
and $W$ is the sum of weights in the neighborhood.
Then, the final radiance $\vec{r}$ and density $\sigma$ are obtained as 
\begin{equation}
    (\vec{r}, \sigma) = F_\psi\left(\vec{h}, \vec{d}\right) 
\end{equation}
based on features $\vec{h}$ and view direction $\vec{d}\in\mathbb{R}^3$.
For computing $\mathcal{N}(\vec{x})$ we created a custom PyTorch extension that implements an efficient, batch-wise voxel grid for ray sample $k$-nn queries in CUDA. The module is based on the Point-NeRF code~\cite{pointnerf} and will be made publicly available with the rest of the code.

Note that the rendering network is purely local, only receiving relative coordinates between ray samples and neural points. Thus, it is only able to learn a local surface model during training and no global, category-level information. The shared features $\mathbf{S}$ are a key ingredient in this formulation. In our studies, we found that they help to train a high-quality category-level neural point renderer and that they represent a density/radiance template of the category. We refer to the supplemental material for further discussion.

\begin{figure*}[h]
  \centering
  \includegraphics[width=1\linewidth]{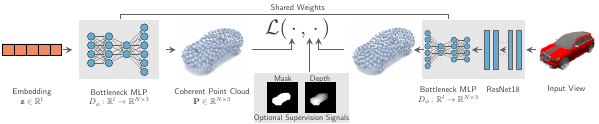}
   \caption{\textbf{Coherent point cloud prediction.} 
   An MLP with bottleneck is used to enforce the point cloud to be constructed as a low-rank deformation of a template (second to last layer output). During training, all trainable parameters (\textcolor{training}{$\blacksquare$}, \textcolor{inference}{$\blacksquare$}) are optimized using ground-truth point cloud supervision. During inference, the embedding $\mathbf{z}$ (\textcolor{inference}{$\blacksquare$}) can be optimized using different supervision signals: a ResNet predicting a point cloud from single image, mask, or depth.}
   \label{fig:point_clouds_method}
   \vspace{-0.2cm}
\end{figure*}

\subsection{Training and Inference}
\label{sec:training_and_inference}
For training and inference, we adopt the autodecoder framework~\cite{deepsdf, deepls}, thus, optimizing embeddings using gradient descent instead of predicting them using an encoder. Recent research has shown that approaches based on such test-time optimization have better capabilities when it comes to accurately representing the given observation~\cite{pointnerf}. 

As input during training, we assume multi-view renderings of a large set of objects, including camera poses for each view. For simplicity, we omit camera poses from the following formulations. Let \mbox{$\mathbf{V} = f_{\theta, \psi}(\mathbf{P}, \mathbf{S}, \mathbf{A}, \mathbf{E})$} denote the function that renders an image $\mathbf{V}$ from a neural point cloud with positions $\mathbf{P}$, embeddings $\mathbf{E}$, attention scores $\mathbf{A}$, and shared features $\mathbf{S}$ from a given camera pose. 

\paragraph{Training} Given a dataset $\{(\mathbf{P}_i, \mathbf{I}_{i,j})\}_{i=j=1}^{K,W}$ of $K$ objects of one category with coherent point clouds $\mathbf{P}_i$, and $W$ renderings $\{\mathbf{I}_{i,j}\}_{j=1}^{W}$ from random views for each object $i$, our training procedure jointly optimizes embeddings $\mathbf{E}_i$ (omitted here for simplicity), attention scores $\mathbf{A}$, shared point features $\mathbf{S}$, and rendering network parameters $\theta, \psi$:
\begin{equation}
    \hat{\mathbf{A}}, \hat{\mathbf{S}}, ´\hat{\theta}, \hat{\psi} =  \argmin_{\mathbf{A}, \mathbf{S}, ´\theta, \psi} \sum_{i,j}\mathcal{L}\left(f_{\theta, \psi}(\mathbf{P}_i, \mathbf{S}, \mathbf{A}, \mathbf{E}_i), \mathbf{I}_{i,j}\right) \textnormal{,}
\end{equation}

where $\mathcal{L}$ is chosen as \emph{Mean Squared Error (MSE)}.
After training, we freeze point renderer parameters $(\hat{\theta},\hat{\psi})$, shared features $\hat{\mathbf{S}}$ and category-level symmetries $\hat{\mathbf{A}}$, which are used to guide gradients to the correct embeddings during test-time optimization. 

\paragraph{Inference} During inference, we assume that the camera pose is known. Given one or multiple views $\{\mathbf{I}_{j}\}_{j=1}^{W}$ and the coherent point cloud $\mathbf{P}$, we can find embeddings $\hat{\mathbf{E}}$ by test-time optimization:

\begin{equation}
    \hat{\mathbf{E}} = \argmin_{\mathbf{E}} \sum_{i}\mathcal{L}\left(f_{\hat{\theta}, \hat{\psi}}(\mathbf{P}, \hat{\mathbf{S}}, \hat{\mathbf{A}}, \mathbf{E}), \mathbf{I}_i\right) \textnormal{.}
\end{equation}
Afterwards, the example can be rendered from other views.

\subsection{Coherent Point Clouds from Single Image}
Until now, coherent point clouds were assumed as given for the sake of simplicity. Still, they also have to be obtained from a single image during test time and are a crucial part of our method. We separate the point cloud prediction from the rest of \meth~such that it can be tackled independently. Note that we do not optimize point positions based on rendering losses. 

The full architecture for point cloud prediction is shown in Figure \ref{fig:point_clouds_method}.
Just like for our rendering branch, we follow the autodecoder framework in order to allow for flexible supervision at test time.
However, unlike the \emph{local} neural point features for capturing fine details, we opt for a \emph{global} representation of the point cloud in form of a single learnable latent code $\vec{z}\in\mathbb{R}^l$.
During training, we optimize these embeddings jointly together with an MLP decoder $D_\phi:\mathbb{R}^l \rightarrow \mathbb{R}^{N\times 3}$ predicting the neural point coordinates in canonical space, given by the normalized orientation and position of ShapeNet objects.
The output of the second to last layer is a low-dimensional bottleneck in order to serve as a low-rank representation
of a high-dimensional point template from $\mathbb{R}^{N\times 3}$~\cite{keytr}.
The global representation together with this low-rank regularization allows us to always obtain coherent point clouds irrespective of the final supervision signal.

Given point clouds $\{\mathcal{P}_i\}_{i=1}^K$  of training examples, we train $D_\phi$ and latent codes $\{\vec{z}_i\}_{i=1}^K$ to minimize the 3D Chamfer Distance (CD):
\begin{equation}
    \mathcal{L}_{CD} = \sum_{i=1}^K\textnormal{CD} (D_\phi(\vec{z}_i), \mathcal{P}_i) \textnormal{.}
\end{equation}
As one form of point cloud supervision during test time, we jointly train a ResNet18~\cite{resnet} encoder with the image, segmentation mask, and camera ray encodings~\cite{3dim} as input to predict the same point cloud after decoding with $D_\phi$, by minimizing the MSE.
At test time, we optimize the latent codes while keeping encoder and decoder fixed.
At first sight, it may seem counterintuitive to bring back an encoder into the autodecoder framework.
However, by parameterizing the latent code directly, we gain flexibility regarding additional supervisory signals.

Two optional sources of additional point cloud supervision at test time are the segmentation mask and a depth map, \eg, captured by an RGB-D camera.
For the mask, we use a 2D Chamfer loss between its pixel coordinates and the 2D projection of the point cloud. 
Depth maps can be utilized by projecting the individual depth samples into the world frame and employing an asymmetric 3D Chamfer loss, i.e., minimizing the distance between each sample and its nearest point in the point cloud.
Note that although occluded points from the perspective of the depth map are not explicitly supervised by this optional loss, they are still optimized implicitly because of our global latent representation.

Besides leveraging these additional forms of data, test-time optimization of the point cloud is directly compatible with multi-view supervision. Given multiple views, the parameterized latent code $\vec{z}$ fuses the point clouds predicted individually for each image by the encoder. The following experimental results demonstrate that our neural point representation can benefit significantly from flexible point cloud supervision at test time.

\label{sec:point_clouds}

\section{Experiments and Results}
\label{sec:experiments}
\begin{table*}[h]
\small
\centering

\begin{tabular}{c|l|ccc|ccc|ccc}
\toprule
&& \multicolumn{3}{c|}{1 Input View} & \multicolumn{3}{c|}{1 Input View (Sym.)} & \multicolumn{3}{c}{2 Input Views} \\ 
Cat. & Method & PSNR$\uparrow$ & SSIM$\uparrow$ & LPIPS$\downarrow$ & PSNR$\uparrow$ & SSIM$\uparrow$ & LPIPS$\downarrow$ & PSNR$\uparrow$ & SSIM$\uparrow$ & LPIPS$\downarrow$ \\ 
\midrule
& SRN~\cite{srn}  & 22.22 & 0.893 & 0.107 & 22.05 & 0.893 & 0.106 & 24.82 & 0.920 & 0.093 \\ 
& PixelNeRF~\cite{pixelnerf} & \underline{23.17} & 0.905 & 0.112 & \underline{22.42} & 0.895 & 0.122 & 25.67 & \underline{0.935} & 0.083 \\ 
& CodeNeRF\footnotemark[1]~\cite{codenerf} & \textbf{23.80} & \underline{0.91}\hphantom{0} & 0.106 & \textemdash\footnotemark[1] & \textemdash\footnotemark[1] & \textemdash\footnotemark[1] & \underline{25.71} & 0.93\hphantom{0} & 0.099 \\ 
& FE-NVS\footnotemark[2]~\cite{fe-nvs} & 22.83 & 0.906 & 0.099 & \textemdash\footnotemark[2] & \textemdash\footnotemark[2] & \textemdash\footnotemark[2] & 24.64 & 0.927 & \underline{0.085} \\ 
& VisionNeRF~\cite{visionnerf} & 22.88 & \underline{0.908} & \underline{0.085} & 22.07 & \underline{0.898} & \underline{0.091} & \textemdash\footnotemark[3] & \textemdash\footnotemark[3] & \textemdash\footnotemark[3] \\ 
\rowcolor{rowhighlight}%
\cellcolor{white} & \textbf{Ours} & 23.00 & \textbf{0.911} & \textbf{0.081} & \textbf{22.67} & \textbf{0.910} & \textbf{0.084} & \textbf{26.69} & \textbf{0.948} & \textbf{0.055} \\
\cmidrule{2-11} 
\parbox[t]{2mm}{\multirow{-8}{*}{\rotatebox[origin=c]{90}{Cars}}} & {\color{grey}\textbf{Ours} + GT PC} & {\color{grey} \textbf{23.69}} & {\color{grey} \textbf{0.920}} & {\color{grey} \textbf{0.077}} & {\color{grey} \textbf{23.31}} & {\color{grey} \textbf{0.916}} & {\color{grey} \textbf{0.080}} & {\color{grey} \textbf{27.29}} & {\color{grey} \textbf{0.953}} &  {\color{grey} \textbf{0.051}} \\ 

\midrule
& SRN~\cite{srn}  & 22.32 & 0.904 & 0.092 & 22.11 & 0.901 & 0.095 & 25.72 & 0.935 & 0.072 \\ 
& PixelNeRF~\cite{pixelnerf} & \underline{23.72} & 0.913 & 0.101 & \underline{23.15} & 0.906 & 0.109 & \underline{26.21} & \underline{0.941} & 0.070 \\
& CodeNeRF\footnotemark[1]~\cite{codenerf} & 23.66 & 0.90\hphantom{0} & 0.136 & \textemdash\footnotemark[1] & \textemdash\footnotemark[1] & \textemdash\footnotemark[1] & 25.63 & 0.91\hphantom{0} & 0.129 \\
& FE-NVS\footnotemark[2]~\cite{fe-nvs} & 23.21 & \underline{0.920} & \textbf{0.077} & \textemdash\footnotemark[2] & \textemdash\footnotemark[2] & \textemdash\footnotemark[2] & 25.25 & 0.940 & \underline{0.065} \\ 
& VisionNeRF~\cite{visionnerf} & \textbf{24.48} & \textbf{0.930} & \textbf{0.077} & \textbf{23.88} & \textbf{0.925} & \textbf{0.085} & \textemdash\footnotemark[3] & \textemdash\footnotemark[3] & \textemdash\footnotemark[3] \\
\rowcolor{rowhighlight}%
\cellcolor{white} & \textbf{Ours} & 23.46 & 0.918 & 0.081 & 23.03 & \underline{0.912} & \underline{0.088} & \textbf{27.36} & \textbf{0.953} & \textbf{0.057} \\ 

\cmidrule{2-11} 

\parbox[t]{2mm}{\multirow{-8}{*}{\rotatebox[origin=c]{90}{Chairs}}} & {\color{grey}\textbf{Ours} + GT PC} & {\color{grey} \textbf{25.80}} & {\color{grey} \textbf{0.940}} & {\color{grey} \textbf{0.071}} & {\color{grey} \textbf{25.24}} & {\color{grey} \textbf{0.934}} & {\color{grey} \textbf{0.077}} & {\color{grey} \textbf{28.43}} & {\color{grey} \textbf{0.961}} &  {\color{grey} \textbf{0.053}} \\
\bottomrule
\end{tabular}

\caption{\textbf{Novel view synthesis on ShapeNet cars and chairs.} 
Our approach achieves state-of-the-art or competitive performance in single-view reconstruction and outperforms previous baselines by a large margin in the two-view setting. While PixelNeRF achieves a higher PSNR with single-view input, it can be seen that their results are blurry (\cf Fig.~\ref{fig:qualitative}). Additionally to the standard setup, (Sym.) provides results on all target views that show mostly the opposite object side to the input view. We show results of our method using point cloud supervision (\textbf{Ours + GT PC}) at test time. These results are only to identify the gap induced by errors in point cloud prediction. The results shown in the rest of the paper use the \textbf{Ours} setup.
}
\label{tab:quantitative}
\vspace{-0.3cm}
\end{table*}

In this section, we describe experiments made with \meth~and present our results. The goal of the experiments is to provide evidence for the following statements: \meth~(1) improves on previous approaches in reconstructing unseen symmetric object parts (c.f.~Section~\ref{sec:single_view_reconstruction}), (2) learns correct self-similarities that respect symmetries of the object category (c.f.~Section~\ref{sec:learned_self-similarities}), (3) provides a meaningful representation that can be used for interpolation (c.f.~Section~\ref{sec:meaningful_representation_space}), (4) is a very efficient approach (c.f.~Section~\ref{sec:efficient_representation}), and (5) is compatible with test-time pose optimization (c.f.~Section~\ref{sec:pose_optimization}).
We provide additional results in the appendix, such as more qualitative results, ablation studies, and an analysis of our point cloud prediction.

\subsection{Experimental Setup}
\label{sec:experimental_setup}
We conduct experiments for category-specific novel view synthesis, given one or two input views.
We make use of the ShapeNet dataset provided by SRN~\cite{srn}, which comprises $3514$ cars and $6591$ chairs split into training, validation, and test sets.
Each training instance has been rendered from $50$ random camera locations on a sphere around the object.
The test sets have $704$ and $1318$ examples, respectively, all with the same $251$ views from an Archimedean spiral and the same lighting as during training.
As done by all baselines, view $64$ and additionally view $104$ are used as input for single- and two-view experiments.
The image resolution is $128\times 128$.
We use PSNR, SSIM, and LPIPS~\cite{lpips} to compare our approach with SRN~\cite{srn}, PixelNeRF~\cite{pixelnerf}, FE-NVS~\cite{fe-nvs}, and VisionNeRF~\cite{visionnerf}.

We use $512$/$4096$ points and $512$/$128$ $32$-dim. embeddings per object.
The shared features are of dimensionality $64$.
After separate pretraining of the point cloud prediction, we train for 0.9/1.2M iterations.
At test time, the instance-specific parameters are optimized for 10k iterations each.
Please refer to the suppl. mat. for additional information.

\begin{figure*}
  \centering
    \begin{subfigure}[t]{0.495\textwidth}
    \includegraphics[width=1\linewidth]{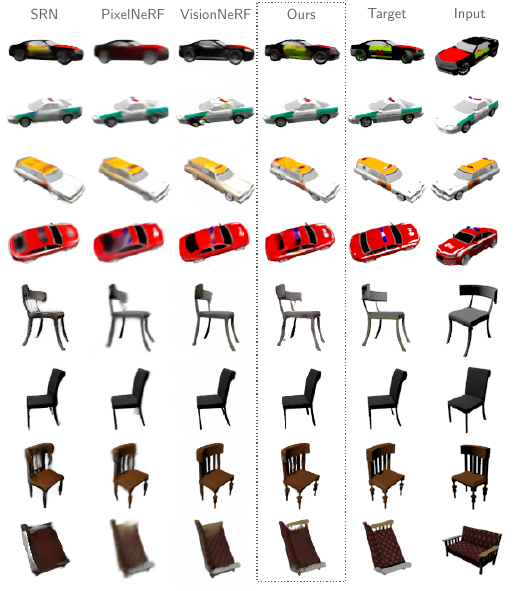}
    \caption{\textbf{Qualitative single-view reconstruction.}}\label{fig:edge_details}
    \label{fig:qualitative}
  \end{subfigure}
  \hfill
  \begin{subfigure}[t]{0.495\textwidth}
    \includegraphics[width=1\linewidth]{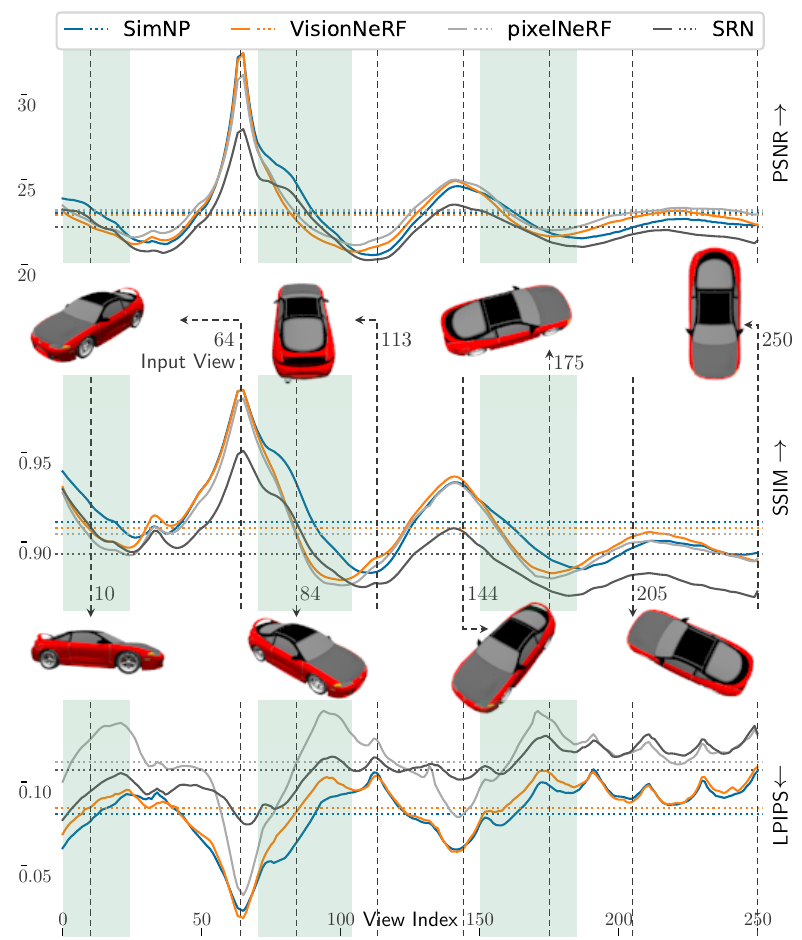}
    \caption{\textbf{Metrics per view.}}\label{fig:noisy_details}
    \label{fig:view_metrics}
  \end{subfigure}
  \caption{\textbf{a)} Our method enables more detailed reconstructions and can better transfer appearance
information to symmetric regions compared to SRN, PixelNeRF, and VisionNeRF. \textbf{b)} We show metric comparisons for each target view in the 251-view SRN test spiral for single-view reconstruction of cars. As visible, our overall better performance can be attributed to views showing regions symmetric to the
input view (green areas and example views 10, 84, 175).
  Also, the related object-level method SRN shows rather flat curves indicating a weak adaptation to observations. This is not the case for \meth. 
  }
  \vspace{-0.2cm}
\end{figure*}

\footnotetext[1]{CodeNeRF~\cite{codenerf} does not provide trained models. Note that the input views differ from the usual setup.}
\footnotetext[2]{FE-NVS~\cite{fe-nvs} does not have public code. Thus, shown numbers are taken from the paper/from authors, where available. SRN, PixelNeRF and VisionNeRF are evaluated using the PixelNeRF calc\_metrics script.}
\footnotetext[3]{VisionNeRF~\cite{visionnerf} not applicable for multiple input views.}

\subsection{Single- and Two-View Reconstruction}
\label{sec:single_view_reconstruction}
Table~\ref{tab:quantitative} summarizes our quantitative results.
We leverage three different test-time optimization setups. In our main setup (\textbf{Ours}), we use the ResNet and the mask for test time optimization of point clouds to ensure a fair comparison with previous approaches. This setup is used for all comparisons and qualitative results shown. Furthermore, we report results using ground-truth point clouds (\textbf{Ours + GT PC}) and provide results with additional ground-truth depth supervision in the supplement. 

In single-view reconstruction of cars, \meth~outperforms the state-of-the-art in SSIM and LPIPS.
We attribute the small PSNR advantage of PixelNeRF to the blurriness of its reconstructions visible in Fig.~\ref{fig:qualitative}, which is confirmed by higher LPIPS. Also, we outperform PixelNeRF in inferring symmetric regions, as we show in columns~\mbox{3-6} in Table~\ref{tab:quantitative} and in Figure~\ref{fig:view_metrics}. 
SRN lacks details due to its global representation.
PixelNeRF's reconstructions are overly-blurry in unseen regions, because pixel-aligned features are shared along the same rays of the input view without respecting the category-level self-similarities.
VisionNeRF is not able to consistently transfer uncommon local color variations to symmetric object regions, which indicates a too strong global prior. In contrast, as visible from Figure~\ref{fig:qualitative} and Figure~\ref{fig:teaser}, our approach correctly transfers local patterns by leveraging learned category-level self-similarities. As visible from Figure~\ref{fig:view_metrics} and columns~\mbox{3-6} in Table~\ref{tab:quantitative}, it significantly outperforms all baselines for views on regions with symmetric counterparts in the input image.

Compared to cars, the chairs category exhibits a larger variety in geometry but less texture details, leading to a smaller advantage through the self-similarity prior. Still, \meth~performs competitively compared to the baselines. Moreover, our strong results under the assumption of a given ground-truth point cloud (last rows in Table~\ref{tab:quantitative}) indicate further potential of our method in combination with an improved point cloud prediction.

In the two-view setting, our approach surpasses previous methods by a large margin on both categories. Here, the benefit comes not only from modeling self-similarities but also from our local point-based representation. The results show that it is better suited for fusing multiple observations than pixel-aligned methods, while at the same time being able to fit details. Looking at the qualitative results in Fig.~\ref{fig:qualitative_2v}, we can observe detailed reconstructions. In Tab.~\ref{tab:2view_settings}, we compare different two view setup against each other. The \emph{same side} setup, which does observe one side of the object with two views, already clearly improves the results.

\begin{table} 
\small
\centering
\begin{tabular}{c|c|ccc}
\toprule
Cat. & View Idx. & PSNR$\uparrow$ & SSIM$\uparrow$ & LPIPS$\downarrow$ \\ 
\midrule
& single view (64) & 23.00 & 0.911 & 0.081 \\
& 2 v. close (63,64) & 23.39 & 0.916 & 0.078 \\  
& 2 v. same side (44,64) & 25.76 & 0.938 & 0.064 \\
\parbox[t]{2mm}{\multirow{-4}{*}{\rotatebox[origin=c]{90}{Cars}}} & 2 v. opposite (64,104) & 26.69 & 0.948 & 0.055 \\
\midrule
& single view (64) & 23.46 & 0.918 & 0.081 \\
& 2 v. close (63,64) & 24.24 & 0.924 & 0.077 \\  
& 2 v. same side (44,64) & 26.50 & 0.944 & 0.064 \\
\parbox[t]{2mm}{\multirow{-4}{*}{\rotatebox[origin=c]{90}{Chairs}}} & 2 v. opposite (64,104) & 27.36 & 0.953 & 0.057 \\
\bottomrule
\end{tabular}
\caption{\textbf{Effect of input views.} 
We evaluate different two view setups, comparing two very close views, 2 views from the same side of the object (approx. 90 degrees apart) and 2 views from opposite sides of the object. It can be seen that even if one side of the object is not seen (same side setup), the results improve strongly with 2 views.
}
\label{tab:2view_settings}
\vspace{-0.2cm}
\end{table}

\begin{figure}
  \centering
  \includegraphics[width=\columnwidth]{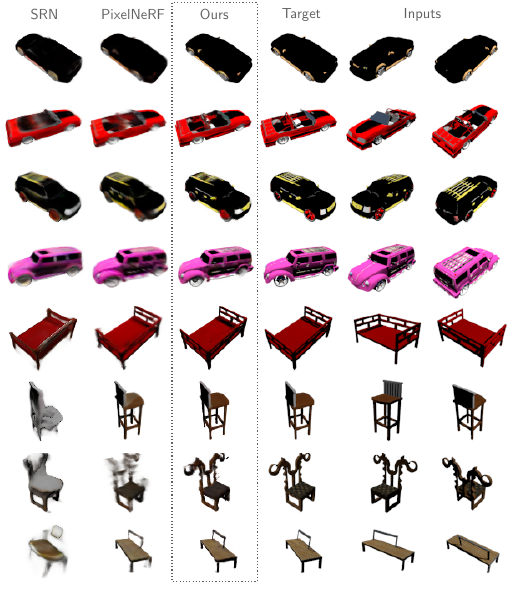}
  \caption{\textbf{Two-view reconstruction.} \meth~learns a high-quality 3D object representation given only two input views.}
  \label{fig:qualitative_2v}
\vspace{-0.2cm}
\end{figure}

\subsection{Learned Self-Similarities}
\label{sec:learned_self-similarities}

As \meth~learns category-level self-similarities explicitly, we can directly inspect them to gain more insights.
For visualizing the attention between neural points and embeddings, we proceed as follows.
The rendering network computes weights between ray samples and neural points (see Eq.~\ref{eq:weights}). By treating these weights just like RGB channels during ray marching, we obtain the influence of each neural point on each pixel. Finally, we multiply the learned category-level attention scores of a single embedding to the respective neural point influence, resulting in the self-similarity visualizations given in Figure~\ref{fig:attention}. We describe the visualization procedure in detail in the supplemental material.
\begin{figure}
  \centering
  \includegraphics[width=\columnwidth]{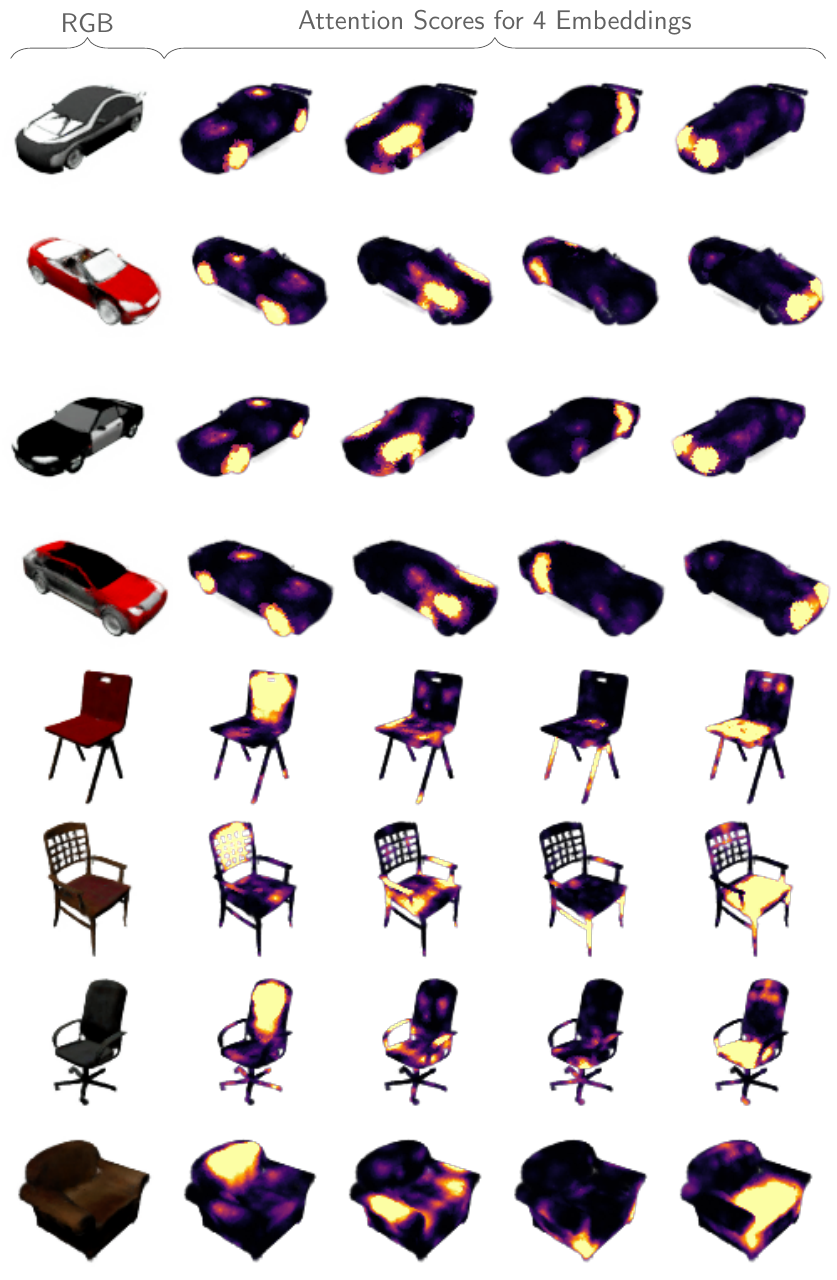}
  \caption{\textbf{Attention visualization.} We render the influence of four different embeddings per category (one per column) on each ray, and visualize them on eight different examples. It becomes evident that embeddings specialize on self-similar local areas, \eg, wheels, or front lights. All embeddings show plane symmetric patterns, showing that our method recovers the symmetry of the category.}
  \label{fig:attention}
\vspace{-0.2cm}
\end{figure}
Although we use as many embeddings as neural points (for cars) such that the attention could learn a one-to-one mapping, the representation learns to share information between similar points.
Further, we observe that the connections for all embeddings are symmetric with respect to a reflection plane, indicating that the method successfully learned a prior about the plane symmetry of each category.

\subsection{Meaningful Representation Space}
\label{sec:meaningful_representation_space}

Figure~\ref{fig:interpolation} provides results for interpolating the point cloud latent code and/or the embeddings of two instances obtained by multi-view fitting.
\meth~learns a meaningful, disentangled representation space allowing for smooth shape and/or appearance transitions from one object to another. This is in contrast to pixel-aligned methods, which are not suited for interpolation. As an example, Fig.~\ref{fig:interpolation_comparison} shows a comparison with VisionNeRF, for which interpolation of the intermediate feature maps results in pixel space interpolation.

\begin{figure}
  \centering
  \includegraphics[width=\columnwidth]{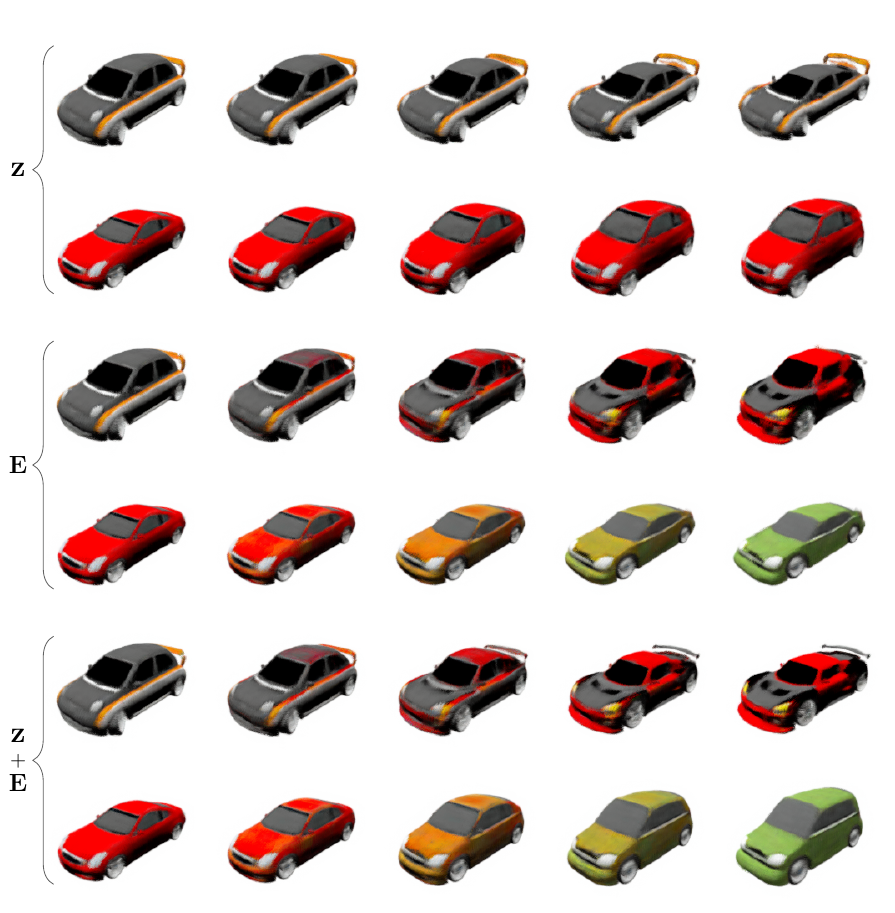}
  \caption{\textbf{Disentangled interpolation.} Our point-based representation disentangles the basic shape from appearance. Point cloud latent code $\vec{z}$ and embeddings $\mathbf{E}$ can be interpolated independently resulting in a smooth shape and/or appearance transition between different objects.}
  \label{fig:interpolation}
\end{figure}

\begin{figure}
  \centering
  \includegraphics[width=\columnwidth]{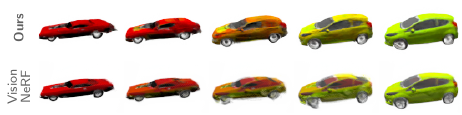}
  \caption{\textbf{Interpolation comparison.} Unlike VisionNeRF, which sticks to interpolation in pixel space for views similar to the input perspective, \meth~enables semantically meaningful interpolation like SRN but with more details.}
  \label{fig:interpolation_comparison}
\end{figure}

\subsection{Efficient Representation}
\label{sec:efficient_representation}
\meth~renders a frame in $59$ms, which is more than an order of magnitude faster than the pixel-aligned methods PixelNeRF ($2.116$s) and VisionNeRF ($2.700$s). We cannot fairly compare against FE-NVS, as no public code is available. The evaluation they provide in the paper is using amortized rendering, which is not comparable, as it only works on many views simultaneously. However, we suspect that their method is also very fast. We train our method on a single A40 GPU for \mbox{$\approx 1.6$ days}. In comparison, SRN and PixelNeRF take $6$ days on an RTX 8000 or Titan RTX, respectively, FE-NVS $2.5$ days on $30$ V100, and VisionNeRF around $5$ days on $16$ A100 GPUs, according to the authors. 
Overall, our method is very efficient with resource-efficient training and fast rendering that shows potential to be applied to larger scale reconstruction in future work. 

\subsection{Pose Optimization}
\label{sec:pose_optimization}
\begin{figure}
  \centering
  \includegraphics[width=\columnwidth]{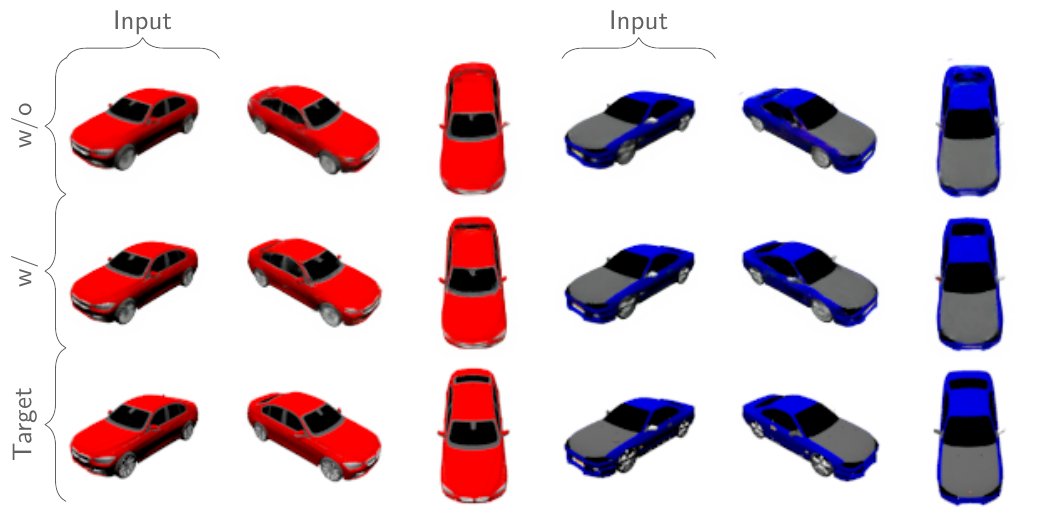}
  \caption{\textbf{Pose Optimization.} \meth~is compatible with camera pose optimization. The figure compares single-view reconstruction results with and without given camera pose.}
  \label{fig:cam_opt}
\end{figure}

To alleviate the assumption of a known camera pose at test time, we investigate optimization of camera parameters. More precisely, for each test example, we initialize eight evenly distributed camera locations on the sphere around the object. The rotation matrix is always defined by the z-axis pointing towards the sphere center and the y-axis pointing upwards.
At test time, we first initialize the point cloud latent using an encoder without ray encodings.
Then, the pose in form of the camera location on the sphere is optimized by minimizing the Chamfer loss between the mask's pixel coordinates and the 2D projection of the point cloud.
Lastly, we further finetune the point cloud with the Chamfer loss and optimize the embeddings for reconstruction of the input image.
We select the pose that leads to the best reconstruction based on LPIPS against the single input image. 
Fig.~\ref{fig:cam_opt} shows a comparison given vs. optimized camera pose.
We achieve competitive performance (0.116 LPIPS) compared with CodeNeRF (0.114) without camera poses.

\section{Conclusion}
We presented \meth, the first object-level representation based on neural radiance fields, utilizing shared attention scores to learn category-specific self-similarities. Our method reaches state-of-the-art quality in single- and two-view reconstruction, while being highly efficient. Overall, it achieves a better \emph{data prior vs. observation trade-off}. The improvements can be contributed to (1) leveraging a local neural point radiance field for the representation of object categories and (2) correctly propagating information between similar regions.
\paragraph{Limitations and Future Work}
The main limitation of \meth~is the assumption of a canonical space with ground-truth point clouds during training, which prohibits direct application on in-the-wild datasets.
Therefore, an improved point cloud prediction in camera frame could be an exciting path for future work.
Furthermore, we believe that it is promising to relax the point identities and to apply our self-similarity priors on a scene level to obtain large-scale, prior-driven reconstruction.

{\small
\bibliographystyle{ieee_fullname}
\bibliography{egbib}

\begin{thebibliography}{10}\itemsep=-1pt

\bibitem{pbgraphics}
Kara-Ali Aliev, Artem Sevastopolsky, Maria Kolos, Dmitry Ulyanov, and Victor
  Lempitsky.
\newblock Neural point-based graphics.
\newblock In {\em European Conhference on Computer Vision (ECCV)}, 2020.

\bibitem{renderdiffusion}
Titas Anciukevicius, Zexiang Xu, Matthew Fisher, Paul Henderson, Hakan Bilen,
  Niloy~J. Mitra, and Paul Guerrero.
\newblock {RenderDiffusion}: Image diffusion for {3D} reconstruction,
  inpainting and generation.
\newblock {\em arXiv pre-print}, 2022.

\bibitem{twoview}
François ARJ, Medioni GG, and Waupotitsch R.
\newblock Mirror symmetry: 2-view stereo geometry.
\newblock In {\em Image and Vision Computing}, 2003.

\bibitem{mipnerf}
Jonathan~T. Barron, Ben Mildenhall, Matthew Tancik, Peter Hedman, Ricardo
  Martin-Brualla, and Pratul~P. Srinivasan.
\newblock {Mip-NeRF}: A multiscale representation for anti-aliasing neural
  radiance fields.
\newblock 2021.

\bibitem{deepls}
Rohan Chabra, Jan~E. Lenssen, Eddy Ilg, Tanner Schmidt, Julian Straub, Steven
  Lovegrove, and Richard Newcombe.
\newblock Deep local shapes: Learning local {SDF} priors for detailed {3D}
  reconstruction.
\newblock In {\em European Conhference on Computer Vision (ECCV)}, 2020.

\bibitem{eg3d}
Eric~R. Chan, Connor~Z. Lin, Matthew~A. Chan, Koki Nagano, Boxiao Pan,
  Shalini~De Mello, Orazio Gallo, Leonidas Guibas, Jonathan Tremblay, Sameh
  Khamis, Tero Karras, and Gordon Wetzstein.
\newblock Efficient geometry-aware {3D} generative adversarial networks.
\newblock In {\em Conference on Computer Vision and Pattern Recognition
  (CVPR)}, 2022.

\bibitem{mvsnerf}
Anpei Chen, Zexiang Xu, Fuqiang Zhao, Xiaoshuai Zhang, Fanbo Xiang, Jingyi Yu,
  and Hao Su.
\newblock {MVSNeRF}: Fast generalizable radiance field reconstruction from
  multi-view stereo.
\newblock {\em Int. Conference on Computer Vision (ICCV)}, 2021.

\bibitem{autosweep}
Xin Chen, Yuwei Li, Xi Luo, Tianjia Shao, Jingyi Yu, Kun Zhou, and Youyi Zheng.
\newblock Autosweep: Recovering 3d editable objects from a single photograph.
\newblock In {\em ACM Trans. Gr.}, 2018.

\bibitem{gsm}
Zhiqin Chen and Hao Zhang.
\newblock Learning implicit fields for generative shape modeling.
\newblock In {\em Conference on Computer Vision and Pattern Recognition
  (CVPR)}, 2019.

\bibitem{stereoradiance}
Julian Chibane, Aayush Bansal, Verica Lazova, and Gerard Pons-Moll.
\newblock Stereo radiance fields ({SRF}): Learning view synthesis from sparse
  views of novel scenes.
\newblock In {\em Conference on Computer Vision and Pattern Recognition
  (CVPR)}. {IEEE}, 2021.

\bibitem{3dr2n2}
Christopher~B Choy, Danfei Xu, JunYoung Gwak, Kevin Chen, and Silvio Savarese.
\newblock {3D-R2N2}: A unified approach for single and multi-view 3d object
  reconstruction.
\newblock In {\em European Conhference on Computer Vision (ECCV)}, 2016.

\bibitem{vit}
Alexey Dosovitskiy, Lucas Beyer, Alexander Kolesnikov, Dirk Weissenborn,
  Xiaohua Zhai, Thomas Unterthiner, Mostafa Dehghani, Matthias Minderer, Georg
  Heigold, Sylvain Gelly, Jakob Uszkoreit, and Neil Houlsby.
\newblock An image is worth 16x16 words: Transformers for image recognition at
  scale.
\newblock In {\em Int. Conference on Learning Representations (ICLR)}, 2021.

\bibitem{pointsetgen}
Haoqiang Fan, Hao Su, and Leonidas Guibas.
\newblock A point set generation network for 3d object reconstruction from a
  single image.
\newblock In {\em Conference on Computer Vision and Pattern Recognition
  (CVPR)}, 2017.

\bibitem{extractstruct}
Roger Fawcett, Andrew Zisserman, and Michael Brady.
\newblock Extracting structure from an affine view of a 3d point set with one
  or two bilateral symmetries.
\newblock In {\em Image and Vision Computing}.

\bibitem{mirrorsym}
Alexandre François, Gérard Medioni, and Roman Waupotitsch.
\newblock Reconstructing mirror symmetric scenes from a single view using
  2-view stereo geometry.
\newblock In {\em ICPR: Proceedings of the 16th International Conference on
  Pattern Recognition}, 2002.

\bibitem{predrepobj}
Rohit Girdhar, David~F. Fouhey, Mikel~D. Rodriguez, and Abhinav~Kumar Gupta.
\newblock Learning a predictable and generative vector representation for
  objects.
\newblock 2016.

\bibitem{fe-nvs}
Pengsheng Guo, Miguel~Angel Bautista, Alex Colburn, Liang Yang, Daniel
  Ulbricht, Joshua~M. Susskind, and Qi Shan.
\newblock Fast and explicit neural view synthesis.
\newblock In {\em Winter Conference on Applications of Computer Vision (WACV)},
  2022.

\bibitem{keytr}
Shreyas Hampali, Sayan~Deb Sarkar, Mahdi Rad, and Vincent Lepetit.
\newblock Keypoint transformer: Solving joint identification in challenging
  hands and object interactions for accurate 3d pose estimation.
\newblock In {\em Conference on Computer Vision and Pattern Recognition
  (CVPR)}, 2022.

\bibitem{resnet}
Kaiming He, Xiangyu Zhang, Shaoqing Ren, and Jian Sun.
\newblock {Deep Residual Learning for Image Recognition}.
\newblock In {\em Conference on Computer Vision and Pattern Recognition
  (CVPR)}, 2016.

\bibitem{affinerecon}
Du~Q. Huynh.
\newblock Affine reconstruction from monocular vision in the presence of a
  symmetry plane.
\newblock In {\em Int. Conference on Computer Vision (ICCV)}, 1999.

\bibitem{snes}
Eldar Insafutdinov, Dylan Campbell, Jo{\~a}o~F Henriques, and Andrea Vedaldi.
\newblock {SNeS}: Learning probably symmetric neural surfaces from incomplete
  data.
\newblock In {\em European Conhference on Computer Vision (ECCV)}, 2022.

\bibitem{codenerf}
Wonbong Jang and Lourdes Agapito.
\newblock Codenerf: Disentangled neural radiance fields for object categories.
\newblock In {\em Int. Conference on Computer Vision (ICCV)}, 2021.

\bibitem{geonerf}
M. Johari, Y. Lepoittevin, and F. Fleuret.
\newblock {GeoNeRF}: Generalizing nerf with geometry priors.
\newblock In {\em Conference on Computer Vision and Pattern Recognition
  (CVPR)}, 2022.

\bibitem{catmesh}
Angjoo Kanazawa, Shubham Tulsiani, Alexei~A. Efros, and Jitendra Malik.
\newblock Learning category-specific mesh reconstruction from image
  collections.
\newblock In {\em European Conhference on Computer Vision (ECCV)}, 2018.

\bibitem{dispointflow}
Roman Klokov, Edmond Boyer, and Jakob Verbeek.
\newblock Discrete point flow networks for efficient point cloud generation.
\newblock In {\em European Conhference on Computer Vision (ECCV)}, 2020.

\bibitem{deformnet}
Andrey Kurenkov, Jingwei Ji, Animesh Garg, Viraj Mehta, JunYoung Gwak, Chris
  Choy, and Silvio Savarese.
\newblock {DeformNet}: Free-form deformation network for {3D} shape
  reconstruction from a single image.
\newblock 2018.

\bibitem{dense3dsym}
Kevin Köser, Christopher Zach, and Marc Pollefeys.
\newblock Dense 3d reconstruction of symmetric scenes from a single image.
\newblock In {\em Proceedings of the 33rd international conference on Pattern
  recognition}, 2011.

\bibitem{symmnerf}
Xingyi Li, Chaoyi Hong, Yiran Wang, Zhiguo Cao, Ke Xian, and Guosheng Lin.
\newblock Symmnerf: Learning to explore symmetry prior for single-view view
  synthesis.
\newblock In {\em Proceedings of the Asian Conference on Computer Vision
  (ACCV)}, pages 1726--1742, December 2022.

\bibitem{visionnerf}
Kai-En Lin, Lin Yen-Chen, Wei-Sheng Lai, Tsung-Yi Lin, Yi-Chang Shih, and Ravi
  Ramamoorthi.
\newblock Vision transformer for {NeRF}-based view synthesis from a single
  input image.
\newblock In {\em Winter Conference on Applications of Computer Vision (WACV)},
  2023.

\bibitem{nsvf}
Lingjie Liu, Jiatao Gu, Kyaw~Zaw Lin, Tat-Seng Chua, and Christian Theobalt.
\newblock Neural sparse voxel fields.
\newblock In {\em Int. Conference on Neural Information Processing Systems
  (NIPS)}, 2020.

\bibitem{onet}
Lars Mescheder, Michael Oechsle, Michael Niemeyer, Sebastian Nowozin, and
  Andreas Geiger.
\newblock Occupancy networks: Learning 3d reconstruction in function space.
\newblock In {\em Conference on Computer Vision and Pattern Recognition
  (CVPR)}, 2019.

\bibitem{nerf}
Ben Mildenhall, Pratul~P. Srinivasan, Matthew Tancik, Jonathan~T. Barron, Ravi
  Ramamoorthi, and Ren Ng.
\newblock {NeRF}: Representing scenes as neural radiance fields for view
  synthesis.
\newblock In {\em European Conhference on Computer Vision (ECCV)}, 2020.

\bibitem{exploitsym}
Dipti~Prasad Mukherjee, Andrew Zisserman, and Michael Brady.
\newblock Shape from symmetry: Detecting and exploiting symmetry in affine
  images.
\newblock In {\em Philosophical Transactions: Physical Sciences and
  Engineering}, 1995.

\bibitem{dvr}
Michael Niemeyer, Lars Mescheder, Michael Oechsle, and Andreas Geiger.
\newblock Differentiable volumetric rendering: Learning implicit 3d
  representations without 3d supervision.
\newblock In {\em Conference on Computer Vision and Pattern Recognition
  (CVPR)}, 2020.

\bibitem{deepsdf}
Jeong~Joon Park, Peter Florence, Julian Straub, Richard Newcombe, and Steven
  Lovegrove.
\newblock {DeepSDF}: Learning continuous signed distance functions for shape
  representation.
\newblock In {\em Conference on Computer Vision and Pattern Recognition
  (CVPR)}, 2019.

\bibitem{seeingglass}
Cody~J. Phillips, Matthieu Lecce, and Kostas Daniilidis.
\newblock Seeing glassware: from edge detection to pose estimation and shape
  recovery.
\newblock In {\em Robotics: Science and Systems}, 2016.

\bibitem{pbgraphics+}
Ruslan Rakhimov, Andrei-Timotei Ardelean, Victor Lempitsky, and Evgeny Burnaev.
\newblock {NPBG++}: Accelerating neural point-based graphics.
\newblock In {\em Conference on Computer Vision and Pattern Recognition
  (CVPR)}, 2022.

\bibitem{schoenberger2016sfm}
Johannes~Lutz Sch\"{o}nberger and Jan-Michael Frahm.
\newblock Structure-from-motion revisited.
\newblock In {\em Conference on Computer Vision and Pattern Recognition
  (CVPR)}, 2016.

\bibitem{schoenberger2016mvs}
Johannes~Lutz Sch\"{o}nberger, Enliang Zheng, Marc Pollefeys, and Jan-Michael
  Frahm.
\newblock Pixelwise view selection for unstructured multi-view stereo.
\newblock In {\em European Conference on Computer Vision (ECCV)}, 2016.

\bibitem{recon3Dmirror}
Sudipta~N. Sinha, Krishnan Ramnath, and Richard Szeliski.
\newblock Detecting and reconstructing 3d mirror symmetric objects.
\newblock In {\em European Conhference on Computer Vision (ECCV)}, 2012.

\bibitem{srn}
Vincent Sitzmann, Michael Zollh{\"o}fer, and Gordon Wetzstein.
\newblock Scene representation networks: Continuous {3D}-structure-aware neural
  scene representations.
\newblock In {\em Int. Conference on Neural Information Processing Systems
  (NIPS)}, 2019.

\bibitem{sv3dlearn}
Maxim Tatarchenko, Stephan~R. Richter, Rene Ranftl, Zhuwen Li, Vladlen Koltun,
  and Thomas Brox.
\newblock What do single-view 3d reconstruction networks learn?
\newblock In {\em Conference on Computer Vision and Pattern Recognition
  (CVPR)}, 2019.

\bibitem{shapefromsym}
Sebastian Thurn and Ben Wegbreit.
\newblock Shape from symmetry.
\newblock In {\em Int. Conference on Computer Vision (ICCV)}, 2005.

\bibitem{posendf}
Garvita Tiwari, Dimitrije Antic, Jan~Eric Lenssen, Nikolaos Sarafianos, Tony
  Tung, and Gerard Pons-Moll.
\newblock Pose-ndf: Modeling human pose manifolds with neural distance fields.
\newblock In {\em European Conference on Computer Vision ({ECCV})}, October
  2022.

\bibitem{grf}
Alex Trevithick and Bo Yang.
\newblock Grf: Learning a general radiance field for 3d scene representation
  and rendering.
\newblock In {\em Int. Conference on Computer Vision (ICCV)}, 2021.

\bibitem{unannotatedimage}
Shubham Tulsiani, Nilesh Kulkarni, and Abhinav Gupta.
\newblock Implicit mesh reconstruction from unannotated image collections.
\newblock In {\em arXiv pre-print}, 2020.

\bibitem{ibrnet}
Qianqian Wang, Zhicheng Wang, Kyle Genova, Pratul Srinivasan, Howard Zhou,
  Jonathan~T. Barron, Ricardo Martin-Brualla, Noah Snavely, and Thomas
  Funkhouser.
\newblock Ibrnet: Learning multi-view image-based rendering.
\newblock In {\em Conference on Computer Vision and Pattern Recognition
  (CVPR)}, 2021.

\bibitem{3dim}
Daniel Watson, William Chan, Ricardo~Martin Brualla, Jonathan Ho, Andrea
  Tagliasacchi, and Mohammad Norouzi.
\newblock Novel view synthesis with diffusion models.
\newblock In {\em Int. Conference on Learning Representations (ICLR)}, 2023.

\bibitem{learninglatent}
Jiajun Wu, Chengkai Zhang, Tianfan Xue, William~T. Freeman, and Joshua~B.
  Tenenbaum.
\newblock Learning a probabilistic latent space of object shapes via 3d
  generative-adversarial modeling.
\newblock In {\em Int. Conference on Neural Information Processing Systems
  (NIPS)}, 2016.

\bibitem{derendering}
Shangzhe Wu, Ameesh Makadia, Jiajun Wu, Noah Snavely, Richard Tucker, and
  Angjoo Kanazawa.
\newblock De-rendering the world's revolutionary artefacts.
\newblock In {\em Conference on Computer Vision and Pattern Recognition
  (CVPR)}, 2021.

\bibitem{unsupsym}
Shangzhe Wu, Christian Rupprecht, and Andrea Vedaldi.
\newblock Unsupervised learning of probably symmetric deformable 3d objects
  from images in the wild.
\newblock In {\em Conference on Computer Vision and Pattern Recognition
  (CVPR)}, 2020.

\bibitem{pointnerf}
Qiangeng Xu, Zexiang Xu, Julien Philip, Sai Bi, Zhixin Shu, Kalyan Sunkavalli,
  and Ulrich Neumann.
\newblock {Point-NeRF}: Point-based neural radiance fields.
\newblock In {\em Conference on Computer Vision and Pattern Recognition
  (CVPR)}, 2022.

\bibitem{idr}
Lior Yariv, Yoni Kasten, Dror Moran, Meirav Galun, Matan Atzmon, Basri Ronen,
  and Yaron Lipman.
\newblock Multiview neural surface reconstruction by disentangling geometry and
  appearance.
\newblock In {\em Int. Conference on Neural Information Processing Systems
  (NIPS)}, 2020.

\bibitem{pixelnerf}
Alex Yu, Vickie Ye, Matthew Tancik, and Angjoo Kanazawa.
\newblock {pixelNeRF}: Neural radiance fields from one or few images.
\newblock In {\em Conference on Computer Vision and Pattern Recognition
  (CVPR)}, 2021.

\bibitem{nerfpp}
Kai Zhang, Gernot Riegler, Noah Snavely, and Vladlen Koltun.
\newblock Nerf++: Analyzing and improving neural radiance fields.
\newblock {\em arXiv pre-print}, 2020.

\bibitem{lpips}
Richard Zhang, Phillip Isola, Alexei~A. Efros, Eli Shechtman, and Oliver Wang.
\newblock The unreasonable effectiveness of deep features as a perceptual
  metric.
\newblock In {\em Conference on Computer Vision and Pattern Recognition
  (CVPR)}, June 2018.

\bibitem{toch}
Keyang Zhou, Bharat~Lal Bhatnagar, Jan~Eric Lenssen, and Gerard Pons-Moll.
\newblock Toch: Spatio-temporal object-to-hand correspondence for motion
  refinement.
\newblock In {\em European Conference on Computer Vision ({ECCV})}. {Springer},
  October 2022.

\end{thebibliography}
}
\setcounter{section}{0}
\renewcommand\thesection{\Alph{section}}
\section*{Appendix}
In this supplementary material, we present ablation studies of our method \textbf{\meth} in Section~\ref{sec:ablation}. Section~\ref{sec:pc_supervision} deals with additional point cloud supervision signals at test time. We make an argument about the quality of the PSNR metric in Section~\ref{sec:effect_of_blur}. Additional visualizations of learned symmetries (including a detailed description of how they are obtained) are shown in Section~\ref{sec:learned_symmetries}. 
Sections~\ref{sec:evaluation} and \ref{sec:architecture} elaborate on details regarding the evaluation and architecture.
Finally, we provide additional qualitative results in Section~\ref{sec:qualitative}.

\section{Ablation Studies}
\label{sec:ablation}

\paragraph{Representing Category-Level Self-Similarities.}
\label{sec:ablation_symmetries}
In Tab.~\ref{tab:ablations}, we present the results of our ablation studies with respect to shared features, attention definition, and number of embeddings.
We can observe that the shared features are essential to train a high-quality category-level neural point renderer. They are further investigated qualitatively in Section~\ref{sec:templates}.
We also test to obtain our matrix $\vec{A}$ via dot products between optimized keys (one per embedding) and queries (one per neural point) instead of directly optimizing $\vec{A}$, which leads to a marginal drop in performance.
As this alternative formulation is more flexible in terms of the number of neural points though, the results indicate potential for an extension of \meth~from object to scene level.
Finally, with an increasing number of embeddings, PSNR and SSIM decrease slightly in turn for improved LPIPS.
We explain this behavior with the effect of blur on the different metrics, which we further investigate in Sec.~\ref{sec:effect_of_blur}.
The smaller the number of embeddings, the smoother are the reconstructions, up to the same number of embeddings as neural points (512). Even more embeddings result in less information sharing and therefore worse generalization to novel views. In total, we can observe that except for existence of shared features, our approach is very robust to changes in these hyperparameters as the quality of the results differs only slightly.

\begin{table}[h]
\centering
\begin{tabular}{ccc|c}
\toprule
Rays & Mask & Aug & CD$\downarrow$ ($\times 10^{-3}$) \\ 
\midrule
\xmark & \xmark & \xmark & 1.2103 \\
\cmark & \xmark & \xmark & 1.1977 \\
\xmark & \cmark & \xmark & 1.1634 \\
\cmark & \cmark & \xmark & 1.1434 \\
\rowcolor{rowhighlight}%
\cmark & \cmark & \cmark & 1.1028 \\
\bottomrule
\end{tabular}
\caption{\textbf{Point cloud prediction ablation.}
The 3D Chamfer distance is computed between the ground-truth point cloud and the one obtained by decoding the ResNet18 output for view $64$ of all test examples.
By using ray encodings~\cite{3dim} (Rays) and the segmentation mask (Mask) as additional inputs for the encoder, we can improve the point cloud prediction. Furthermore, random color jitter and grayscale augmentations (Aug) result in better generalization.
}
\label{tab:pc_ablations}
\end{table}

\begin{table}[h]
\centering
\begin{tabular}{l|ccc}
\toprule
Method & PSNR$\uparrow$ & SSIM$\uparrow$ & LPIPS$\downarrow$ \\ 
\midrule
PixelNeRF~\cite{pixelnerf} & \underline{23.17} & 0.905 & 0.112 \\
\rowcolor{rowhighlight}%
\textbf{Ours} & 23.00 & \underline{0.911} & \textbf{0.081} \\
\textbf{Ours} + Blur & \textbf{23.31} & \textbf{0.913} & \underline{0.092} \\
\bottomrule
\end{tabular}
\caption{\textbf{Effect of blur.}
By applying a Gaussian blur on our rendered images, we can boost our PSNR results to outperform the strongest baseline with respect to this metric. However, this comes with worse results in LPIPS.
}
\label{tab:effect_of_blur}
\end{table}

\begin{table}[h]
\centering

\begin{tabular}{l|ccc}
\toprule
Supervision & PSNR$\uparrow$ & SSIM$\uparrow$ & LPIPS$\downarrow$ \\ 
\midrule
Mask & 23.00 & 0.911 & 0.081 \\ 
Mask + Depth & 23.28 & 0.915 & 0.079 \\
Point Cloud & 23.69 & 0.920 & 0.077 \\ 
\bottomrule
\end{tabular}

\caption{\textbf{Point cloud supervision on ShapeNet cars.}
Utilizing a depth map can partly bridge the gap between our purely 2D point cloud supervision and the use of ground-truth point clouds.
}
\label{tab:pc_supervision}
\end{table}

\begin{table*}
\centering
\begin{tabular}{ccr|ccc|ccc|ccc}
\toprule
\multicolumn{3}{c|}{Configuration} & \multicolumn{3}{c|}{1 Input View} & \multicolumn{3}{c|}{1 Input View (Sym.)} & \multicolumn{3}{c}{2 Input Views} \\ 
S & A & \multicolumn{1}{c|}{$M$} & PSNR$\uparrow$ & SSIM$\uparrow$ & LPIPS$\downarrow$ & PSNR$\uparrow$ & SSIM$\uparrow$ & LPIPS$\downarrow$ & PSNR$\uparrow$ & SSIM$\uparrow$ & LPIPS$\downarrow$ \\ 
\midrule
\xmark & \cmark & 512 & 22.37 & 0.898 & 0.106 & 22.02 & 0.894 & 0.114 & 25.66 & 0.937 & 0.069 \\
\midrule
\cmark & \xmark & 512 & 22.76 & 0.908 & 0.084 & 22.47 & 0.905 & 0.086 & 26.38 & 0.945 & 0.057 \\
\midrule
\cmark & \cmark & 128 & \textbf{23.09} & \textbf{0.913} & 0.083 & \textbf{22.81} & \textbf{0.911} & \textbf{0.084} & \textbf{26.92} & \textbf{0.949} & 0.058 \\
\cmark & \cmark & 256 & 23.06 & 0.912 & 0.082 & 22.76 & 0.910 & \textbf{0.084} & 26.84 & \textbf{0.949} & 0.055 \\
\rowcolor{rowhighlight}%
\cmark & \cmark & 512 & 23.00 & 0.911 & \textbf{0.081} & 22.67 & 0.908 & \textbf{0.084} & 26.67 & 0.948 & \textbf{0.053} \\
\cmark & \cmark & 1024 & 22.81 & 0.909 & 0.083 & 22.47 & 0.905 & 0.087 & 26.37 & 0.947 & 0.054 \\
\bottomrule
\end{tabular}
\caption{\textbf{Ablation studies on ShapeNet cars.}
S represents if shared features $\mathbf{S}$ are being used, A represents direct parameterization of the attention map $\mathbf{A}$ (\cmark) vs. the common calculation of attention with (shared) keys and queries (\xmark), and $M$ is the number of embeddings. The gray row shows the configuration from the main paper.
}
\label{tab:ablations}
\end{table*}

\paragraph{Coherent Point Cloud Prediction.}
\label{sec:ablation_pc}
We ablate the encoder used for point cloud supervision at test time with respect to additional input data and data augmentations during training. The results are shown in Tab.~\ref{tab:pc_ablations}.
The ray encodings~\cite{3dim} as well as the segmentation mask individually decrease the 3D Chamfer distance between the ground-truth point clouds and the ones predicted for view $64$ of each test example.
Given the ray encodings, the encoder is less likely to confuse the pose of the object, \eg, in case of almost symmetric front and back sides of cars.
Due to the lighting used for rendering the dataset, we have observed that some white cars fade into the white background.
Therefore, we attribute the improvements gained by leveraging the segmentation mask as input to such examples.
Finally, we employ color jitter and grayscale data augmentations during training to enhance generalization.

\section{Additional Point Cloud Supervision}
\label{sec:pc_supervision}
By leveraging the autodecoder framework, we allow for flexible supervision at test time.
Table~\ref{tab:pc_supervision} compares different forms of point cloud supervision.
\meth~can effectively utilize additional depth maps or ground truth point clouds.

\section{Effect of Blur}
\label{sec:effect_of_blur}

To support our claim that the state-of-the-art single-view PSNR results are due to the metric favoring blurry reconstructions, we postprocess our test predictions by applying a Gaussian filter with standard deviation $0.6$.
Table~\ref{tab:effect_of_blur} shows that simply blurring our renderings is enough to raise the PSNR above the one of PixelNeRF~\cite{pixelnerf}.
Interestingly, SSIM is also affected positively, in contrast to LPIPS, which gets worse.
We conclude that the two standard image quality metrics are rather biased towards blurry images such that we suggest to focus more on perceptual metrics like LPIPS.

\section{Category-Level Template}
\label{sec:templates}

\begin{figure*}[h]
  \centering
  \includegraphics[width=\textwidth]{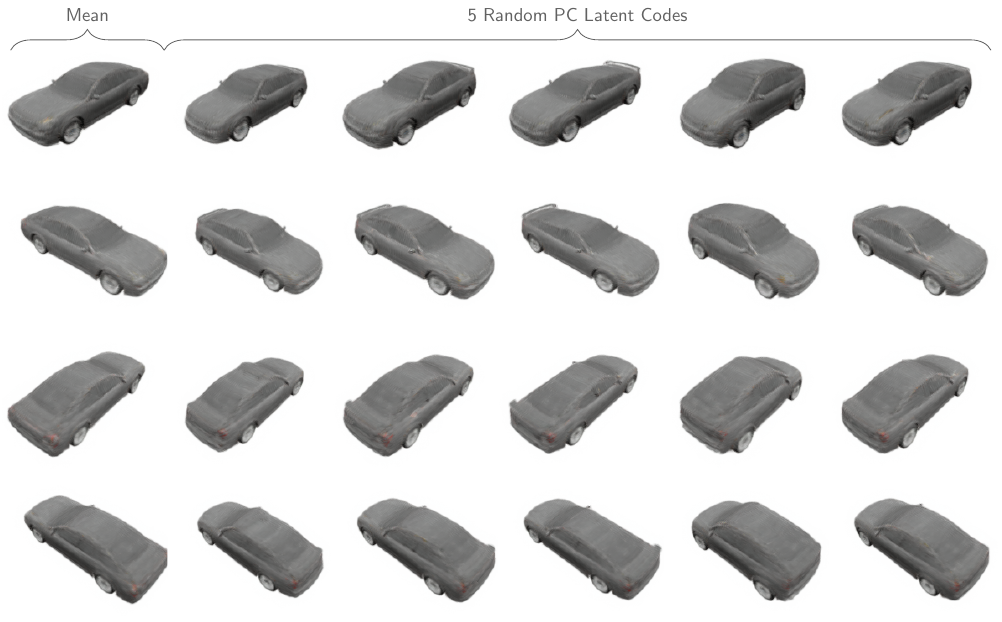}
  \caption{\textbf{Learned car template.} By rendering all-zero embeddings $\vec{E}$, we visualize the template learned by the shared features $\vec{S}$. The left column shows results for a zero point cloud latent code $\vec{z}$, whereas the remaining ones are obtained by sampling random vectors for point clouds.}
  \label{fig:templates}
\end{figure*}

In order to investigate the effect of the shared features, we set the instance-specific embeddings to zero.
Figure~\ref{fig:templates} shows the templates learned by the shared features $\vec{S}$.
Besides the general shape of a car for a given point cloud including details like side mirrors, the shared features also encode common textures like wheels, windows, and lights.
Furthermore, by decoding random point cloud latent codes $\vec{z}$, we always obtain plausible point clouds indicating that point latent plus shared components learn a deformable category-level template, which can be filled with individual details by fitting embeddings $\vec{E}$ to observations.

\section{Learned Symmetries}
\label{sec:learned_symmetries}

We visualize the attention scores for seven more embeddings in Fig.~\ref{fig:attention}.
To evaluate a pixel's radiance $\vec{c}\in\mathbb{R}^3$, the usual volume rendering formulation proposed by NeRF accumulates the radiance for $K$ sample points $\{\vec{x}_i\in\mathbb{R}^3\}_{i=1}^K$ along the ray though the pixel as:
\begin{equation}
    \label{eq:volume_rendering}
    \vec{c} = \sum_{i=1}^K\tau_i(1-\exp(-\sigma_i\Delta_i))\vec{r}_i  \textnormal{,}
\end{equation}
with
\begin{align}
    \tau_i &= \exp\left(-\sum_{j=1}^{i-1}\sigma_j\Delta_j\right) \\
    \Delta_i &= \norm{\vec{x}_{i}-\vec{x}_{i-1}}_2 \textnormal{,}
\end{align}
where $\sigma_i$ and $\vec{r}_i$ are the density and radiance of sample $\vec{x}_i$.
In order to obtain the influence of each neural point on the pixel, we simply replace the radiance $\vec{r}_i$ in Eq.~\ref{eq:volume_rendering} with the normalized inverse distances $\vec{w}_i\in\mathbb{R}^N$ between the sample point and each neural point neighbor:
\begin{equation}
    \vec{w}_i[j]=
    \begin{cases}
        \frac{w(\vec{x}_i,\vec{p}_j)}{W},   & \text{if } j\in\mathcal{N}(\vec{x}_i)\\
        0,                                  & \text{otherwise}\textnormal{,}
    \end{cases}
\end{equation}
where $\mathcal{N}$ is the $k$-nearest neural point function, $\vec{p}_j$ the coordinates of the $j$-th neural point, $w$ the inverse point distance, and $W$ the sum of these weights in the neighborhood, as defined in Section 3.2 of the paper.
Once we have rendered the neural point weights $\vec{c}\in\mathbb{R}^N$ for each pixel, the influence $\vec{i}\in\mathbb{R}^M$ of each embedding can be obtained by multiplying the learned attention scores:
\begin{equation}
    \vec{i} = \textnormal{softmax}(\mathbf{A})\cdot \vec{c}\textnormal{.}
\end{equation}

\section{Evaluation Details}
\label{sec:evaluation}
\paragraph{Evaluation of Render Time.} Table~1 of the paper provides time measurements for rendering a single view.
For a fair comparison, we separated the actual rendering functionality from all preceding inference steps in the code provided by the authors. For example, for VisionNeRF and PixelNeRF we only count the ray casting, feature sampling, and rendering MLPs as rendering, not the inference of feature maps.
We evaluate each method on all $251$ views of five random test examples and average the timings per view. None of the methods use any form of radiance caching or amortized rendering but render each view individually.
The experiments were performed on a single RTX 8000 GPU.

\paragraph{Sym. Setup for Symmetric Views.} Besides the render time, Table~1 of the paper also presents results for single-view reconstruction of views that show mostly the object side opposite to the input view.
To be more precise, we choose the view index intervals 0-33, 74-112, and 152-191.
These are all views with the camera being on the right side of the car up to a certain height, as the input view shows the object from the front left.
Note that this subset also contains views of the rear, which are more challenging for our method because of missing similarities to observed areas.

\section{Architecture Details}
\label{sec:architecture}
The architecture is composed of the point cloud prediction network, the attention representing the category-level symmetries, and the rendering network.
For point cloud prediction, we use a four-layer MLP with hidden dimensions $256$, $128$, $64$, and ReLU activation function.
As input, it gets the latent code $\vec{z}\in\mathbb{R}^l$ with $l=512$.
The final layer outputs the coordinates of all $N=512$ neural points inside a cube of side length $2$ using the hyperbolic tangent.

The rendering network consists of the kernel $K_\theta$ and density and radiance function $F_\psi$.
$K_\theta$ is implemented as a five-layer MLP with output dimension $256$.
$F_\psi$ consists of two separate branches: another five-layer MLP for radiance prediction and a two-layer MLP for the density.
All MLPs of the rendering network use $256$ as the number of hidden dimensions and LeakyReLU as non-linearity.

\section{Additional Qualitative Results}
\label{sec:qualitative}

We present results for coherent point cloud prediction in Figure~\ref{fig:point_clouds}. These results are obtained using the \textbf{Ours} setup from the main paper. The point colors encode point identity over multiple subjects. It can be seen that the resulting point clouds behave \emph{coherent} such that individual points represent the same object parts over multiple instances.

Further, we present additional qualitative results for single-view reconstruction in Figure~\ref{fig:qualitative}, two-view reconstruction in Figure~\ref{fig:qualitative_2v}, and interpolation in Figure~\ref{fig:interpolation}.

\begin{figure*}[h]
  \centering
  \includegraphics[width=\textwidth]{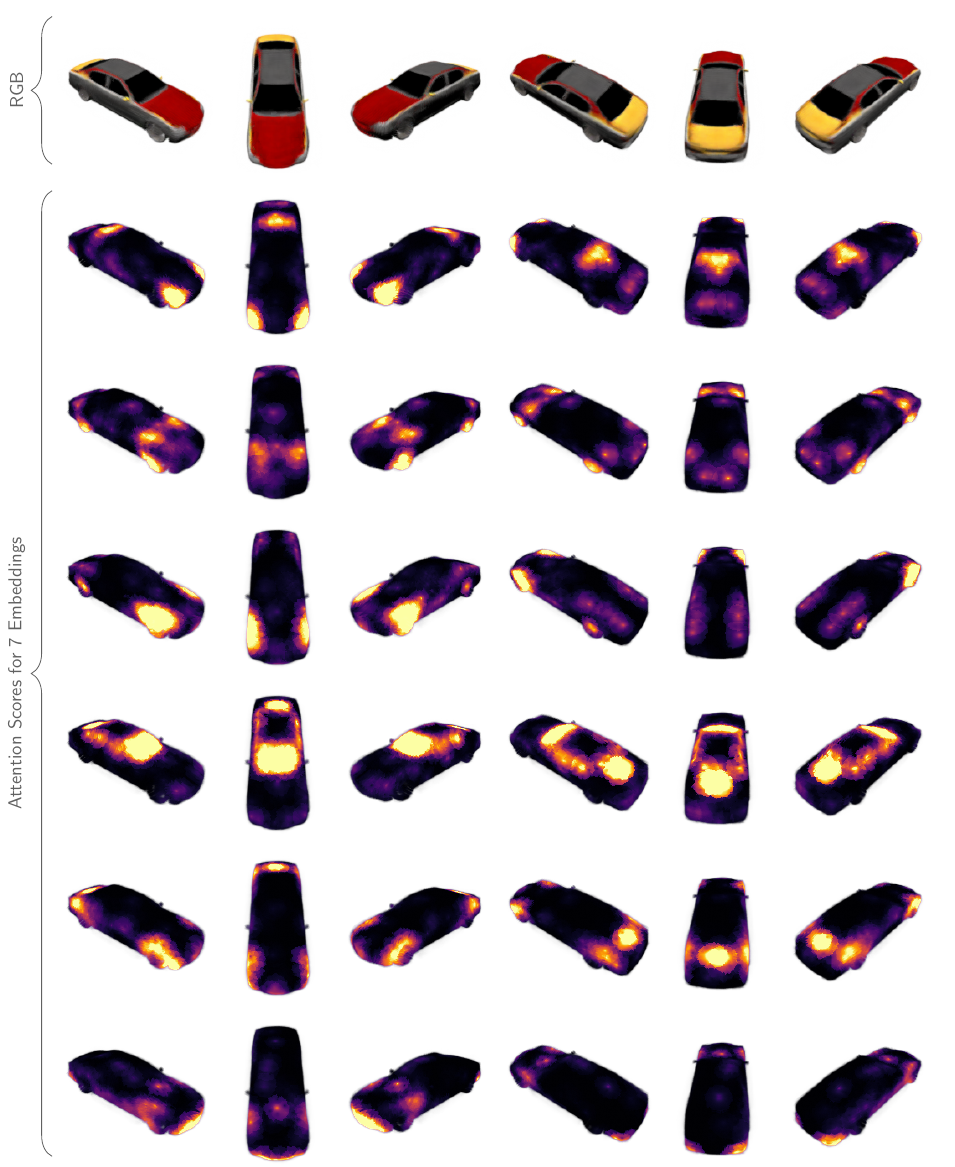}
  \caption{\textbf{Attention visualization.} We render the influence of seven different embeddings (one per row) on each ray.}
  \label{fig:attention}
\end{figure*}

\begin{figure*}[h]
  \centering
  \includegraphics[width=0.85\textwidth]{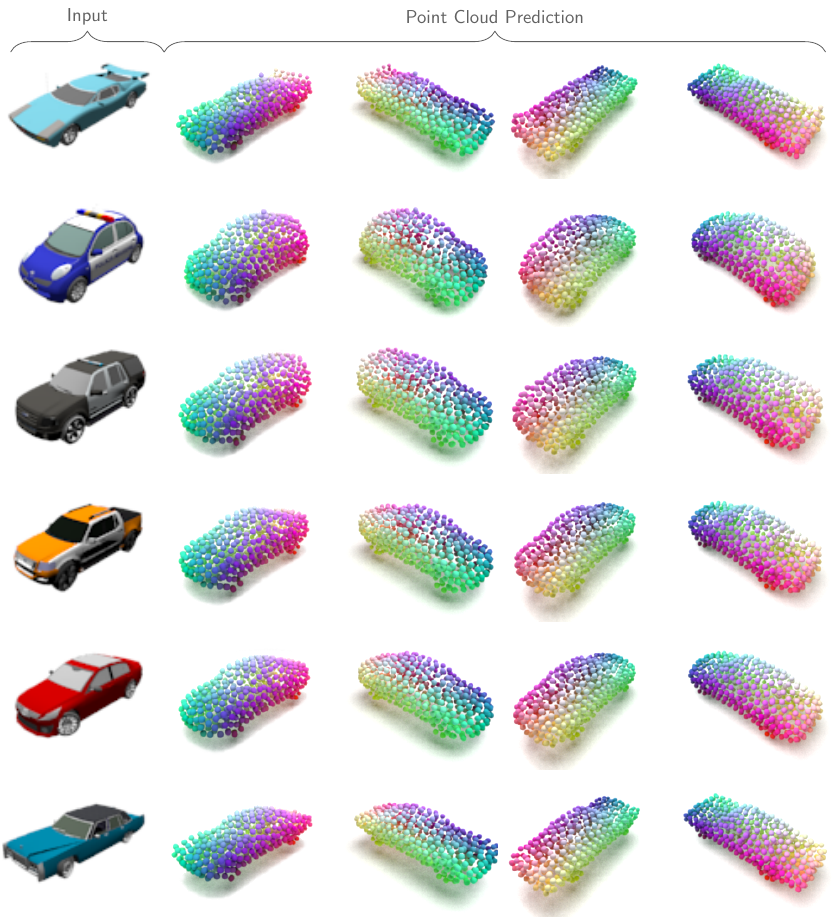}
  \caption{\textbf{Coherent point cloud prediction.} The point color encodes the point identity given by the order of the output tensor that is predicted by the point MLP. It can be seen that these identities behave coherent over multiple instances. This allows the formulation of shared features $\mathbf{S}$. Also, these coherent identities provide correspondences between instances. }
  \label{fig:point_clouds}
\end{figure*}

\begin{figure*}[h]
  \centering
  \includegraphics[width=0.65\textwidth]{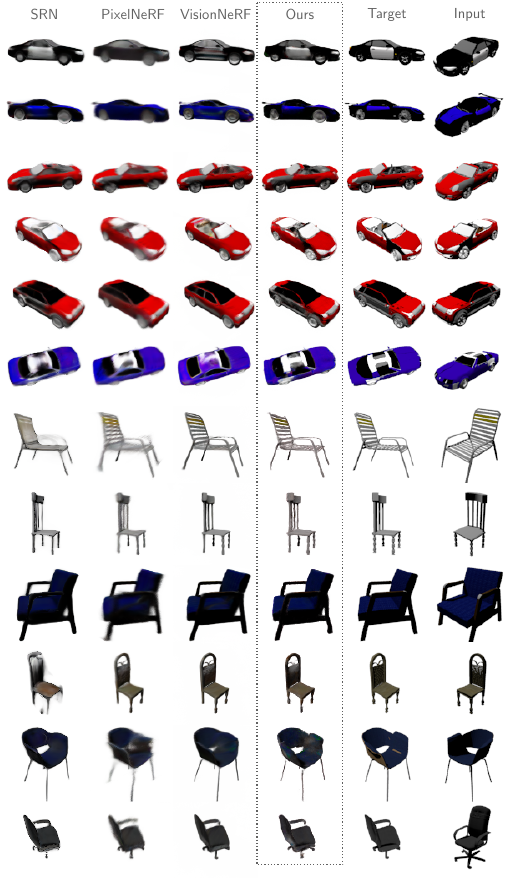}
  \caption{\textbf{Single-view reconstruction.} Additional qualitative results that show that our method is better in replicating details on the symmetric side of the object.}
  \label{fig:qualitative}
\end{figure*}

\begin{figure*}[h]
  \centering
  \includegraphics[width=0.7\textwidth]{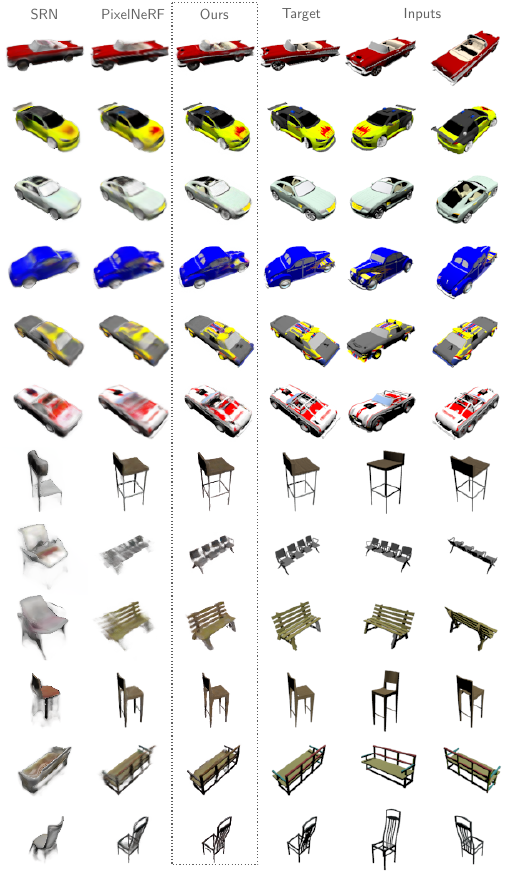}
  \caption{\textbf{Two-view reconstruction.} Additional qualitative results that show our method is better in representing highly detailed objects from just two views.}
  \label{fig:qualitative_2v}
\end{figure*}

\begin{figure*}[h]
  \centering
  \includegraphics[width=0.7\textwidth]{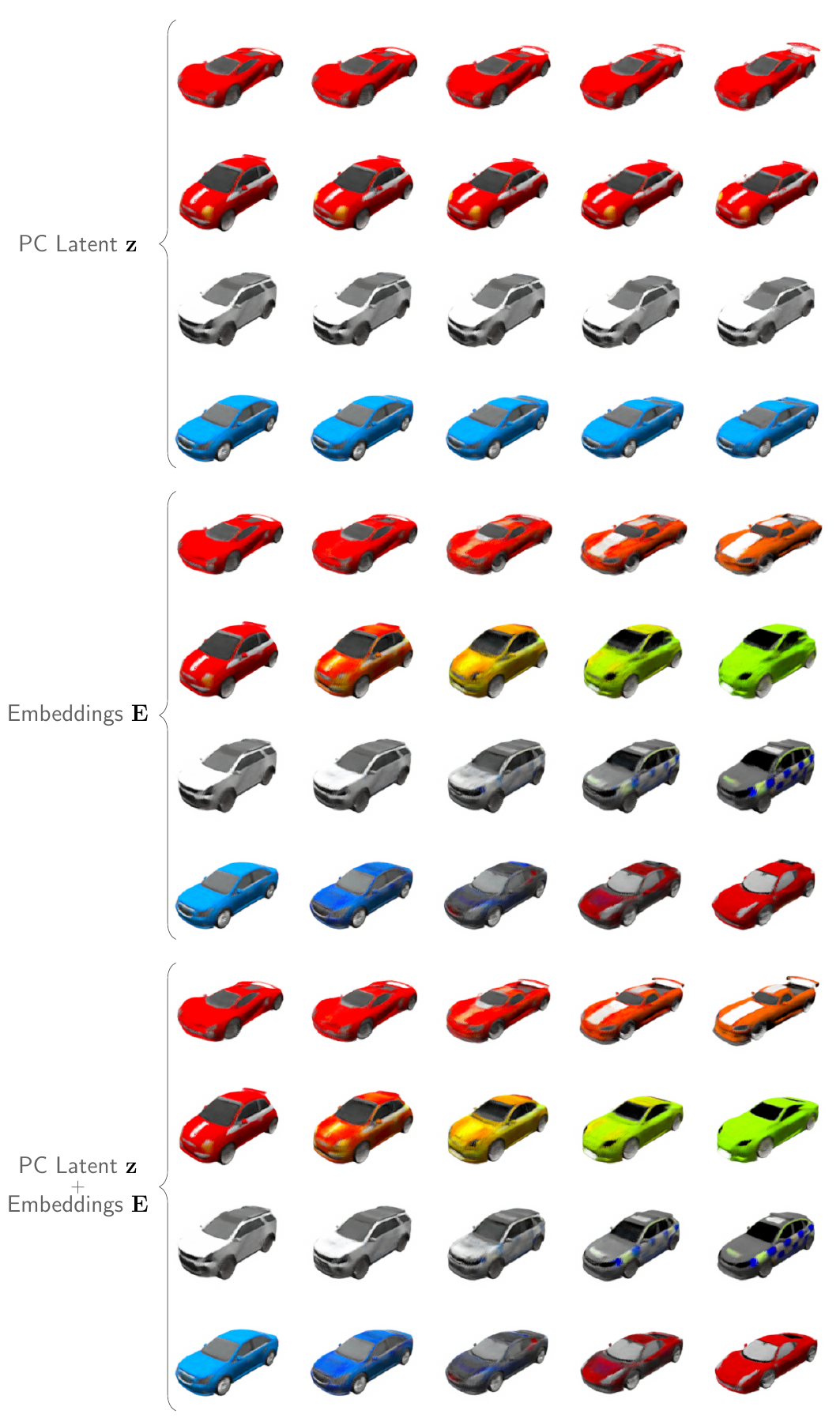}
  \caption{\textbf{Disentangled interpolation.} Additional qualitative results that highlight the ability of interpolating embeddings, point clouds and both together.}
  \label{fig:interpolation}
\end{figure*}

\end{document}